\documentclass[]{bytedance_seed}

\usepackage[toc,page,header]{appendix}

\usepackage{minitoc}
\usepackage{algpseudocode}
\usepackage{algorithm}
\usepackage{amsmath}
\usepackage{amssymb}
\usepackage{mathtools}
\usepackage{bm}
\usepackage{dsfont}
\usepackage{stmaryrd}
\usepackage{systeme}
\usepackage{scalerel}

\usepackage[table]{xcolor}
\usepackage{graphicx}
\usepackage{tikz}
\usetikzlibrary{tikzmark, calc}
\usepackage{float}
\usepackage{stfloats}

\usepackage{booktabs}
\usepackage{multirow}
\usepackage{multicol}
\usepackage{makecell}
\usepackage{adjustbox}

\usepackage{paralist}
\usepackage{outline}
\usepackage{lipsum}
\usepackage{soul}
\usepackage{siunitx}
\usepackage{xspace}
\usepackage{url}
\usepackage{cite}
\usepackage{wrapfig}

\usepackage{algpseudocode}
\usepackage{listings}

\usepackage[font={small}]{caption} 
\usepackage{subcaption} 

\usepackage{titletoc}
\usepackage{tocloft}
\usepackage[nolist]{acronym}

\usepackage{balance}

\usepackage{cuted}
\usepackage{esvect}
\usepackage{stackengine}

\def\ours{VLingNav\xspace}
\DeclareUnicodeCharacter{FF0C}{,}

\newcommand{\eg}{\textit{e}.\textit{g}.}
\newcommand{\ie}{\textit{i}.\textit{e}.}
\newcommand{\etal}{et al.}

\definecolor{myblue}{HTML}{dbe8f5}
\definecolor{mygreen}{HTML}{009900}

\definecolor{BestColor}{HTML}{ffffff}
\definecolor{SecondColor}{HTML}{ffffff}

\title{VLingNav: Embodied Navigation with Adaptive Reasoning and Visual-Assisted Linguistic Memory}

\author{
  Shaoan Wang$^{1,2,*}$, 
  Yuanfei Luo$^{1,*}$,
  Xingyu Chen$^{2,3,\dagger}$,
  Aocheng Luo$^{2}$,\\ 
  Dongyue Li$^{2}$, 
  Chang Liu$^{2}$,
  Sheng Chen$^{1,\ddagger}$, 
  Yangang Zhang$^{1}$, 
  Junzhi Yu$^{2,\dagger}$
}

\affiliation[1]{ByteDance Seed}
\affiliation[2]{Peking University}
\affiliation[3]{Zhongguancun Academy}

\contribution[*]{Co-first authors}
\contribution[\dagger]{Corresponding authors}
\contribution[\ddagger]{Project lead}

\abstract{
Vision-Language-Action (VLA) models have shown promising potential in embodied navigation by unifying perception and planning while inheriting the strong generalization abilities of large Vision-Language Models (VLMs). However, most existing VLA models rely on reactive mappings directly from observations to actions, lacking the explicit reasoning capabilities and persistent memory required for complex, long-horizon navigation tasks. To address these challenges, we propose \ours, a VLA model for embodied navigation grounded in linguistic-driven cognition. First, inspired by the dual-process theory of human cognition, we introduce an adaptive chain-of-thought (AdaCoT) mechanism, which dynamically triggers explicit reasoning only when necessary, enabling the agent to fluidly switch between fast, intuitive execution and slow, deliberate planning. Second, to handle long-horizon spatial dependencies, we develop a visual-assisted linguistic memory module (VLingMem) that constructs a persistent, cross-modal semantic memory, enabling the agent to recall past observations to prevent repetitive exploration and infer movement trends for dynamic environments.
For training, we construct Nav-AdaCoT-2.9M, the largest embodied navigation dataset with reasoning annotations to date, enriched with adaptive CoT annotations that induce a reasoning paradigm capable of adjusting both when to think and what to think about. Moreover, we incorporate an online expert-guided reinforcement learning stage, enabling the model to surpass pure imitation learning and to acquire more robust, self-explored navigation behaviors. Extensive experiments demonstrate that \ours achieves state-of-the-art performance across a wide range of embodied navigation benchmarks. Notably, \ours transfers to real-world robotic platforms in a zero-shot manner, successfully executing practical navigation tasks, including previously unseen and untrained tasks, and demonstrating strong cross-domain and cross-task generalization.
}

\date{\today}
\correspondence{\email{wangshaoan@stu.pku.edu.cn}, \email{luoyuanfei@bytedance.com}}

\checkdata[Project Page]{\url{https://wsakobe.github.io/VLingNav-web/}}

\begin{document}
\maketitle


\begin{figure}[!t]
    \centering
    \includegraphics[width=1\linewidth]{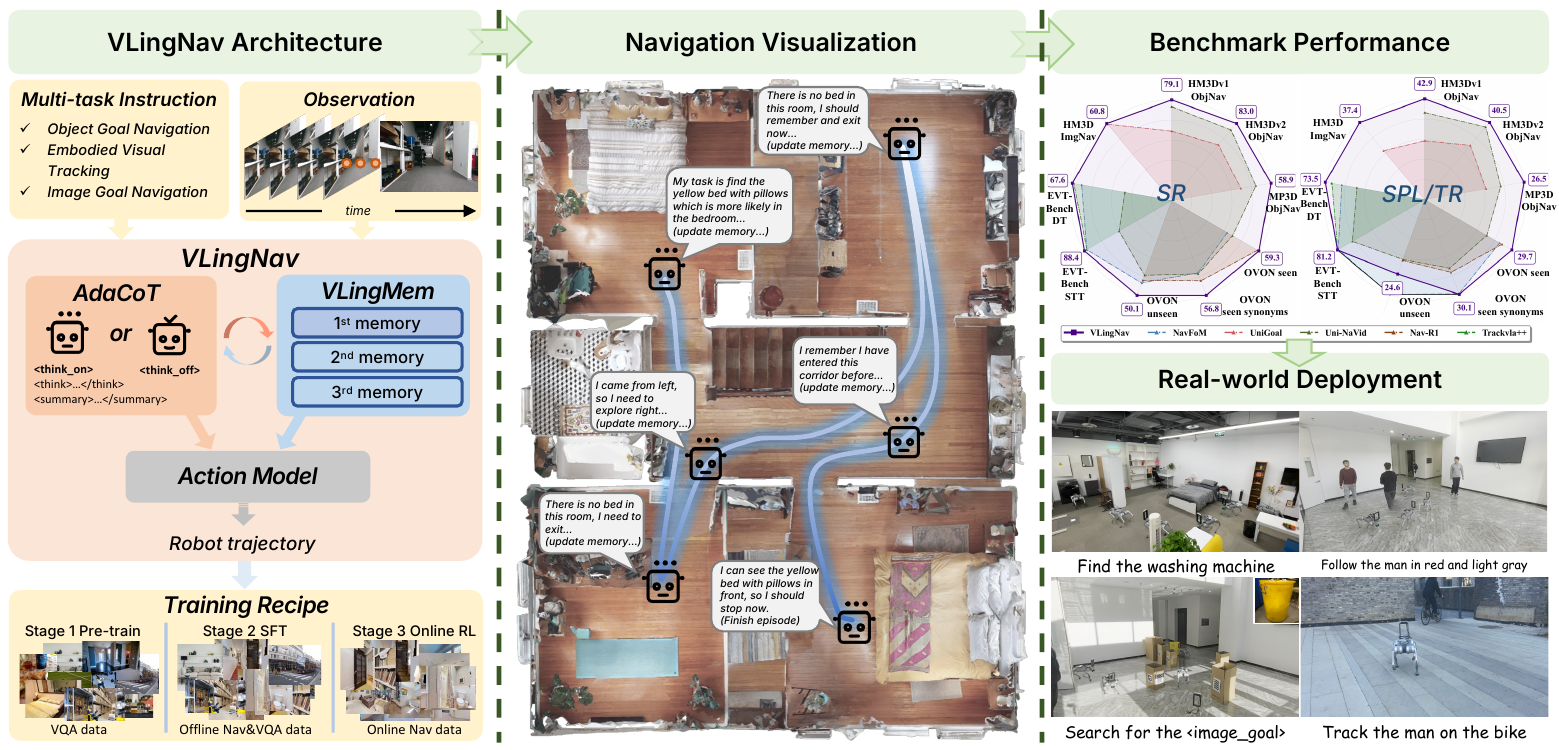}
    \caption{\textbf{Overview of \ours.} \ours is a VLA model enhanced with adaptive CoT reasoning and visual-assisted linguistic memory. This architecture allows the model to leverage historical visual and linguistic memory, achieving SOTA results on several embodied navigation benchmarks. Furthermore, \ours can be deployed zero-shot on real-world robots to perform diverse and complex navigation tasks.}
    \label{fig:teaser}
\end{figure}

\section{Introduction}
Embodied navigation~\citep{wu2024embodied} is a fundamental capability for intelligent robots, enabling purposeful movement through previously unseen, structurally complex environments in response to human instructions. As robots are increasingly deployed in open-world settings from household service scenarios to industrial inspection, navigation systems must deliver accurate perception and decision-making while robustly generalizing to novel scenes and tasks.
Traditional modular approaches~\citep{chang2023goat,yokoyama2024vlfm,chen2025astra} decompose navigation into submodules (e.g., perception, mapping, planning) by leveraging mature techniques such as visual foundation models~\citep{ravi2024sam,liu2023grounding}, SLAM~\citep{qin2018vins,xu2022fast}, and path-planning algorithms~\citep{karaman2011anytime}. However, these pipelines require manually defined module interfaces; over-reliance on hand-crafted rules compromises robustness and induces error accumulation, limiting adaptability in dynamically complex environments. 
Recent advances in large-scale vision–language–action (VLA) models have made compelling progress toward this goal. By unifying multimodal scene understanding with language-conditioned action generation, VLA-based agents substantially improve the adaptability and expressiveness of embodied navigation systems.

Despite this progress, current VLA models are reactive systems, often lacking the explicit reasoning mechanisms, memory structures, and interpretability that are important for reliable real-world deployment. 
Most existing models operate under a fixed inference budget, producing actions with a predetermined amount of computation, and therefore cannot increase deliberation when faced with ambiguity.
In addition, these models often lack persistent semantic memory, relying solely on limited context windows. Without a mechanism to retain historical context, agents struggle to track their progress over extended trajectories, resulting in redundant exploration, looping behaviors, and poor adaptation to dynamic changes in the environment.

Addressing these limitations requires rethinking VLA architectures from a linguistic perspective, moving beyond passive perception-action mapping toward active reasoning, memory construction, and interpretable decision-making. Motivated by principles from cognitive science and human problem solving, we argue that effective embodied navigation demands two missing capabilities:
1) adaptive reasoning, enabling the agent to adjust the granularity of its internal deliberation according to task complexity; and
2) linguistically grounded long-term memory, providing stable cross-modal semantics that support consistent and context-aware navigation behavior.

Furthermore, most current VLA training paradigms rely on supervised fine-tuning (SFT) via imitation learning. However, this approach often limits generalization, preventing models from performing beyond expert demonstrations. While post-training paradigms rooted in reinforcement learning (RL) have proven effective for enhancing LLMs and VLMs on complex tasks~\citep{guo2025deepseek,shao2024deepseekmath,feng2025video}, their application in embodied navigation remains preliminary~\citep{gao2025octonav,liu2025nav,zhang2025activevln}. Notably, existing efforts typically focus on autoregressive RL in discrete space, leaving the exploration of RL for refining continuous control policies an open area for further investigation.

In this work, we present \ours, a linguistic-driven VLA framework designed to endow embodied agents with cognitive abilities through two core components. First, inspired by the fast-and-slow thinking paradigm, we introduce an Adaptive Chain-of-Thought (AdaCoT) mechanism. AdaCoT dynamically triggers explicit reasoning only when necessary, allowing the agent to efficiently switch between fast reactive execution and deliberate planning depending on the situation. Second, to handle long-term spatial dependencies, we develop a Visual-assisted Linguistic Memory module (VLingMem). By constructing a persistent cross-modal memory, VLingMem enables the agent to recall past observations to prevent repetitive exploration and infer movement trends for dynamic tasks, thereby ensuring coherent decision-making over extended interactions.

To support the training of such cognitively enriched VLA models, we construct Nav-AdaCoT-2.9M, the largest embodied navigation dataset with reasoning annotations to date, incorporating adaptive CoT annotations that teach the model when to think and what to think about. Beyond imitation learning, we further employ online expert-guided RL for post-training, enabling \ours to acquire self-improving navigation behaviors that surpass the limitations of supervised demonstrations.

Extensive experiments across diverse embodied navigation benchmarks show that \ours achieves state-of-the-art performance, outperforming existing VLA-based agents in both success rate and efficiency metrics. Notably, \ours transfers to real-world robots in a zero-shot manner, successfully executing novel navigation tasks in the real world without any additional fine-tuning. These results highlight the strong generalization ability of linguistic-driven cognition and demonstrate the promise of integrating adaptive reasoning and persistent memory into VLA models for embodied navigation. The contributions are as follows: 

\begin{itemize}
\item We propose \ours, a novel framework integrating Adaptive Chain-of-Thought (AdaCoT) and Visual-Assisted Linguistic Memory (VLingMem). AdaCoT enables the agent to dynamically switch between fast execution and slow deliberation based on task complexity, while VLingMem eliminates redundant exploration and infers movement trends through persistent cross-modal storage.

\item We construct Nav-AdaCoT-2.9M, the largest embodied navigation dataset with reasoning annotations to date, enriched with adaptive CoT annotations that induce flexible reasoning patterns. We further introduce an online expert-guided RL post-training stage, empowering the model to surpass the limitations of imitation learning and acquire more robust, self-optimized navigation behaviors.

\item We conduct extensive experiments across standard embodied navigation benchmarks, demonstrating that \ours achieves state-of-the-art performance, with significant gains in long-horizon reasoning and success rate. Moreover, \ours exhibits remarkable zero-shot transfer to real-world robot platforms, successfully executing unseen tasks and illustrating strong cross-domain and cross-task generalization.

\end{itemize}

\section{Related Works}
\subsection{Embodied Navigation Models}
As a core task in robotics, navigation has long attracted significant attention from robotics researchers~\citep{desouza2002vision}. With the rise of embodied AI in recent years, robot navigation has gradually shifted from classical point-to-point navigation~\citep{kavraki2002probabilistic} to more intelligent embodied navigation. Embodied navigation includes subtasks such as vision-language navigation (VLN)~\citep{zhang2024navid,zhang2024uni,wei2025streamvln,cheng2024navila,zeng2025janusvln}, object goal navigation (ObjectNav)~\citep{ramrakhya2022habitat,yin2024sg,yadav2023offline}, image goal navigation (ImageNav)~\citep{krantz2023navigating,yadav2023ovrl,yin2025unigoal}, and embodied visual tracking (EVT)~\citep{wang2025trackvla,zhong2024empowering,liu2025trackvla++}, emphasizing that robots follow natural language instructions to perceive, reason, and plan in unseen environments. 

Embodied navigation methods can be broadly categorized into modular and end-to-end approaches. The modular paradigm relies on well-established components such as off-the-shelf large models~\citep{zhou2023navgpt,zhou2025navgpt}, SLAM~\citep{qin2018vins,campos2021orb}, vision foundation models~\citep{zhai2023siglip,radford2021clip}, and planning algorithms~\citep{karaman2011anytime}. It decomposes the navigation task into distinct modules (\eg, perception, localization, planning) and aligns them via manually defined interfaces. This design yields high interpretability and strong zero-shot transfer~\citep{yin2025unigoal}. However, integrating multiple modules inevitably incurs information loss~\citep{luo2018end}; moreover, tight coupling across modules increases system fragility~\citep{luo2018end}.
End-to-end approaches leverage data-driven learning to directly map sensor inputs to robot actions~\citep{yadav2023ovrl,yokoyama2024hm3d,zeng2024poliformer}. By removing manually designed interfaces and mitigating information loss, these methods have achieved notable progress~\citep{ramrakhya2023pirlnav,ramrakhya2022habitat}. However, they exhibit limited generalization and can produce abnormal actions under out-of-distribution conditions. With the rapid advancement of large models in recent years, an increasing number of studies have adopted pre-trained VLMs as the backbone to enhance generalization, environmental perception, and spatial understanding.

NaVid~\citep{zhang2024navid} represents the first embodied navigation VLA model. It designs a video-based VLM and finetunes on VLN datasets, demonstrating robust generalization capabilities. However, its inference time increases significantly with longer video streams, making real-world deployment challenging.
Building upon NaVid, Uni-NaVid~\citep{zhang2024uni} introduces a video-stream compression mechanism to control the number of visual tokens. Moreover, Uni-NaVid extends the model to multiple categories of embodied navigation tasks, achieving state-of-the-art performance across diverse benchmarks. Similarly, NaVILA~\citep{cheng2024navila} and StreamVLN~\citep{wei2025streamvln} adopt similar architectures; they further incorporate large-scale open-world navigation data and leverage KV cache to jointly improve both generalization and inference speed.
JanusVLN~\citep{zeng2025janusvln} enhances 3D understanding by fusing spatial features produced by VGGT~\citep{wang2025vggt}, thereby exhibiting strong instruction-following performance. Notably, all the aforementioned works represent robot actions as discrete tokens. This simplification leads to inefficient action quality and weak adaptability in dynamic scenarios.
To address this limitation, TrackVLA~\citep{wang2025trackvla} designs an anchor-based diffusion policy that directly outputs the robot’s motion trajectory, substantially improving both action quality and efficiency. NavFoM~\citep{zhang2025embodied} further extends the model by introducing TVI tokens, enabling inputs from cross-embodiment navigation data.

Nevertheless, existing navigation VLA models rely solely on action labels for finetuning and thus fail to exploit the inherent reasoning capabilities of VLMs~\citep{bai2025qwen2,feng2025video}. In addition, they maintain history only through implicit visual features, without explicit memory, which ultimately prevents fully unlocking the potential of the VLM backbone.

\subsection{Embodied Chain-of-Thought}
With the chain-of-thought significantly enhancing the performance of LLM and VLM on complex tasks~\citep{wei2022chain,wang2024chain,huang2025thinkact}, several studies have attempted to extend this paradigm to embodied tasks. By explicitly outputting the reasoning process before a robot executes actions, the inherent reasoning capabilities of the VLM can be better leveraged. This approach aims to enhance the model’s competencies in task decomposition, environmental perception, and decision-making, ultimately improving the accuracy and quality of the actions generated by the model, as well as its generalization ability and performance in real-world scenarios. Embodied-CoT~\citep{zawalski2024robotic} first utilizes structured textual instructions enriched with spatial localization information. CoT-VLA~\citep{zhao2025cot} and VPP~\citep{hu2024video,zhang2024up} integrate reasoning via future image prediction. $\pi_{0.5}$~\citep{black2025pi_} performs task decomposition and reasoning through text. \mbox{ChatVLA-2}~\citep{zhouchatvla} enhances the model's performance in complex visual reasoning tasks by introducing additional open-world visual reasoning pre-training data. ThinkAct~\citep{huang2025thinkact} designs a dual-system framework that bridges high-level reasoning with low-level action. However, the aforementioned methods are limited to tabletop manipulation tasks and have not been extended to navigation in open spaces. OctoNav~\citep{gao2025octonav} improves the model's performance in navigation tasks and enhances interpretability by executing CoT at fixed frequency. However, the requirement for manual configuration of the CoT frequency impedes the full exploitation of CoT’s potential. Aux-Think~\citep{wang2025think} constructs a VLN dataset with CoT labeling, and experiments show that using CoT as an auxiliary task during training enhances the model's navigation performance, while excessive reasoning affects the model's efficiency and performance. $NavA^3$~\citep{zhang2025nava} adopts GPT-4o as the reasoning-VLM for task decomposition and 3D spatial localization, but it suffers from long reasoning latency, making it difficult to deploy on real robots.

In contrast to previous work, we propose an adaptive thinking strategy that balances reasoning efficiency with navigation capability.

\subsection{Memory in VLA Models}
For long-horizon embodied tasks, VLA models must possess robust memory capabilities. RoboFlamingo~\citep{li2023vision}, for instance, compresses vision–language representations into latent tokens and propagates them through a Long Short-Term Memory (LSTM) network. However, the resulting latent representations are relatively coarse-grained, leading to a significant loss of fine-grained perceptual history.
In contrast, MemoryVLA~\citep{shi2025memoryvla} integrates high-level cognitive semantics and fine-grained perceptual details within a unified memory framework, enabling effective temporal modeling for long-horizon manipulation tasks. However, it only employs a single implicit cognitive token to serve as the semantic memory, failing to fully leverage the reasoning capabilities of LLM.
On the navigation side, video-based VLA models~\citep{zhang2024navid,zhang2024uni,cheng2024navila,wei2025streamvln,wang2025trackvla,liu2025trackvla++} commonly encode historical image observations as inputs to provide implicit visual memory. However, such implicit memory can hinder learning to focus on key regions, and semantic information is further degraded as visual features are repeatedly compressed.
Finally, Mem2Ego~\citep{zhang2025mem2ego} and MapNav~\citep{zhang-etal-2025-mapnav} incorporate global map information into VLA models as memory components. Yet current VLM backbones lack native support for map-format inputs, and the representation design of maps for VLAs remains under-explored.

Compared with latent-, vision-, or map-based memories, language memory is better aligned with the VLA framework, thanks to large-scale language pretraining. Therefore, we design the memory module from a linguistic perspective and use visual features as auxiliary signals.

\subsection{Post-training for VLA Models}
Reinforcement Learning enhances the exploration capability of large models, unlocks their reasoning potential, and shows promise for mitigating issues such as covariate shift and causal confusion induced by imitation learning. Notably, OctoNav~\citep{gao2025octonav}, VLN-R1~\citep{qi2025vln}, and Nav-R1~\citep{liu2025nav} have convergently integrated GRPO~\citep{shao2024deepseekmath} into navigation VLA models, enabling the simultaneous optimization of CoT outputs and actions. 

\begin{figure*}[!t]
\centering
\includegraphics[width=1\linewidth]{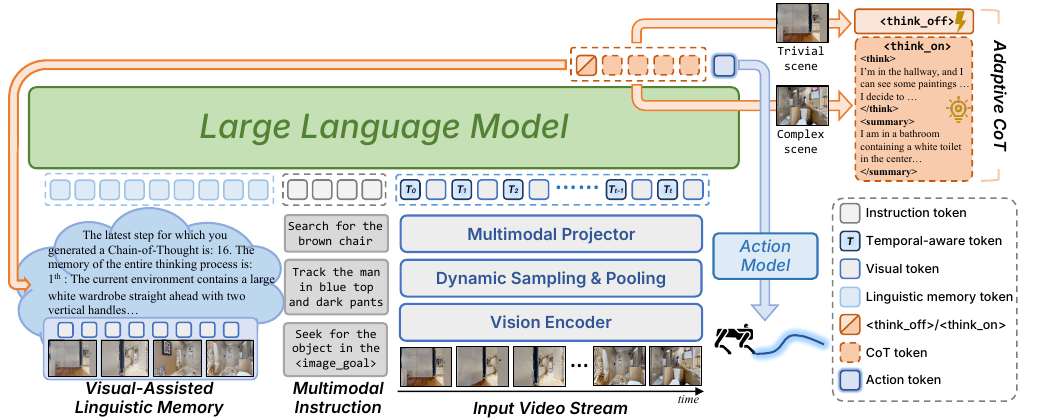}
\caption{\textbf{The overall framework of \ours.} The framework takes video streams and multimodal instruction as input to produce robot action for navigation with tailored linguistic designs. AdaCoT can adaptively generate linguistic thinking according to its observation, while VLingMem summarizes CoT cues with key visual features for globally informed decision-making.}
\label{fig:pipeline}
\end{figure*}

Recent advances in large reasoning models (\eg, \mbox{DeepSeek-R1}~\citep{guo2025deepseek}) show that RL can drive remarkable progress even when relying solely on outcome-based rewards. Several studies have also attempted to leverage outcome-based rewards for the RL post-training of VLA models. For instance, SimpleVLA-RL~\citep{li2025simplevla} pioneers outcome-based rewards for the RL post-training of OpenVLA-OFT~\citep{kim2025fine}, achieving a substantial improvement in success rate on the manipulation benchmarks. ActiveVLN~\citep{zhang2025activevln}, caches all historical actions and states into model tokens and leverages GRPO to implement outcome-based RL through this mechanism. 

The aforementioned work remains confined to autoregressive action outputs, failing to support more advanced continuous action prediction. Recently, ReinFlow~\citep{zhang2025reinflow} addresses this by formulating flow matching as an MDP, enabling RL training via PPO~\citep{schulman2017proximal} or GRPO.

Existing VLA-RL frameworks either adopt discrete autoregressive action with limited policy space or continual flow-based action with slow inference speed. We adopt an MLP-based continuous action to overcome the above drawbacks. In addition, we introduce prior expert knowledge into the RL framework to improve online learning efficiency and performance.
\section{Methodology}

\subsection{Navigation Task Definitions}
Embodied navigation tasks can be defined as follows: a mobile robot is provided with an instruction $\mathcal{I}$ and a sequence of visual observations $\mathcal{O}_{1:t} \in \mathbb{R}^{W \times H \times 3}$, captured by the egocentric camera mounted on the robot at each time step $\{1,...,t\}$. Given these observations and the instruction, the policy model $\pi$ is required to output the next action $a_t \in \mathbb{A}=\{v, \omega\}$ for the robot. The robot accomplishes the navigation task by executing the action predicted by the model, which can be formulated as $a_t = \pi(\mathcal{I}, \mathcal{O}_{1:t})$. \ours is capable of performing multiple embodied navigation tasks, including ObjectNav, EVT, and ImageNav. ObjectNav requires the robot to explore unseen environments given a textual description of an object category and to locate an object that matches the specified goal. EVT focuses on identifying the correct target described by textual instructions in dynamic, crowded scenarios and maintaining continuous tracking of the moving target. ImageNav is analogous to ObjectNav, with the key difference that the goal is specified by an image rather than text. Similarly, the robot must explore unseen environments and find the location corresponding to the image goal.

\subsection{\ours Overview}
\ours extends a video-based VLM, specifically LLaVA-Video-7B~\citep{zhang2024video}, and integrates an action model to enable simultaneous text token generation and trajectory planning. For text token prediction, the model follows a conventional autoregressive paradigm. For trajectory planning, the action model conditions on the VLM backbone’s outputs to predict a motion trajectory $\tau = \{a_1, a_2, \dots\, a_n\}$, where $n$ is the trajectory horizon and each $a \in \mathbb{R}^3 = (x, y, \theta)$ denotes a waypoint that encapsulates both position and orientation.

\subsection{\ours Architecture}

\subsubsection{Observation Encoding}

For the video-based VLA model, the number of image frames grows over time during online inference. This substantially increases computational burden, making it difficult to ensure inference efficiency when deploying on real robots. Moreover, for low-speed mobile robots, adjacent egocentric frames captured at high FPS contain substantial redundant visual information.
Prior studies explore two main strategies to mitigate this issue. One merges visual tokens from historical frames to reduce redundancy among adjacent frames~\citep{zhang2024uni,Bolya2022TokenMY}; however, this operation often distorts original semantic features and introduces additional computation. The other uniformly samples the video stream to reduce frame count~\citep{cheng2024navila}, which inevitably causes delayed and inaccurate decisions due to insufficient short-term observations at low sampling rates. 

To address the limitations of these two approaches, we propose a dynamic FPS sampling strategy. Inspired by the Ebbinghaus forgetting curve~\citep{ebbinghaus2013image}, historical frames are sampled according to their time intervals relative to the current frame. Specifically, older historical frames, regarded as long-term memory, are sampled at a lower rate to simulate the forgetting process. In contrast, recent historical frames, considered as short-term memory, are sampled at a guaranteed higher rate. The relationship between the sampling rate and the time interval approximately satisfies the following:

\begin{equation}
    f_{s}(i) = f_{s}^{max}e^{-\frac{\Delta T}{s}}
\end{equation}
where $f_{s}$ denotes the sampling rate, $f_{s}^{max}$ represents the maximum sampling rate, $\Delta T = t - i$ stands for the time interval from latest frame $t$ to frame $i$, and $s$ signifies the stability of memory. Through this approach, we can control the number of input image tokens while selectively preserving more important images.

After sampling the input visual observations, we need to encode and map the visual observations into the latent space of the VLM backbone. Following LLaVA-Video, we employ a pre-trained vision encoder (SigLIP-400M~\citep{zhai2023siglip}), to encode the input egocentric video stream $\mathcal{O}_{1:t} = \{\mathbf{o}_1, \cdots, \mathbf{o}_t\}$ of the robot. This encoding process yields visual features $\mathbf{V}_{1:t} \in \mathbb{R}^{N \times C}$, where $N$ stands for the number of image patches ($N=729$) and $C$ denotes the embedding dimension ($C=1152$). To efficiently summarize historical visual information, we process past observations using a grid pooling strategy. This approach downsamples the feature maps of historical observations, enabling the model to capture high-level semantic features while effectively controlling computational costs. Similar to dynamic FPS, we also determine the downsampling ratio for grid pooling based on time intervals. The specific operation is defined as follows:
\begin{align}
    g(i) &= \lfloor e^{-\frac{\Delta T}{g}} \rfloor \\
    \mathbf{V}'_{t_i} &= \mathcal{G}(\mathbf{V}_{t_i}, g(i))
\end{align}
where $\mathbf{V}'_{t_i}$ is the $i$-th visual feature after grid pooling operation $\mathcal{G(\cdot)}$ with stride $g(i)$.

Furthermore, to eliminate the temporal inconsistency in the video stream caused by dynamic FPS sampling, we incorporate timestamp information for each frame within the visual observations. Specifically, a temporal-aware indicator token $E^T(\cdot) \in \mathbb{R}^C$ is introduced prior to each frame, which can reflect the time interval between a given historical visual observation and the current observation. By encoding timestamp information using Rotary Position Embedding (RoPE)~\citep{su2024roformer}, $E^T$ enables the model to perceive the absolute time interval between different historical frames and the current frame. It can be expressed as follows:

\begin{equation}
    E^T(\Delta T) = E^T_{base} + RoPE(\Delta T)
\end{equation}

For the projection of visual features, we follow the well-established framework of VLMs~\citep{liu2023llava}. Specifically, a cross-modality projector based on a two-layer Multi-Layer Perceptron (MLP) $\mathcal{P}(\cdot)$ is employed to map the visual features $\mathbf{V}$ into the latent space of the VLM, yielding the projection result as $\mathbf{E}^V_t = \mathcal{P}(\mathbf{V}_t^{'})$, where $\mathbf{E}^V_t$ represents the projected visual token.

\subsubsection{Adaptive CoT \& Visual-Assisted Linguistic Memory}
As illustrated in Fig.~\ref{fig:pipeline}, we concatenate the visual tokens $\mathbf{E}_t^V$ with the language tokens $\mathbf{E}^I$ and the temporal-aware indicator tokens $\mathbf{E}^T$ to form the input sequence of the VLM. To balance the model's inference performance and efficiency, we train the model using a large scale high-quality adaptive CoT data (detailed in Sec.~\ref{sec:data}) endowing it with the ability to autonomously decide whether to perform CoT reasoning for a given input. Specifically, for the current input, the VLM first predicts a CoT indicator token (\texttt{<think\_on>} or \texttt{<think\_off>}). Upon outputting \texttt{<think\_on>}, the model generates the specific content of CoT in an autoregressive manner, which consists of two components:

\begin{itemize}
    \item The reasoning content, enclosed within \texttt{<think>} and \texttt{</think>} tokens. This content includes perception of the visual observation, task decomposition and analysis, assessment of whether the current location has been visited, and determination of the next action.
    \item The environmental summary of the current observation, enclosed within \texttt{<summary>} and \texttt{</summary>} tokens. This summary is incorporated into subsequent inputs as linguistic memory.
\end{itemize}

\subsubsection{Action Model}
To transfer the reasoning and decision-making knowledge of the VLM backbone into the robot-specific action space, we integrate an MLP-based action model $\mathcal{A}_{\theta}(\cdot)$ into \ours. Specifically, the hidden state vector $\textbf{h}_t^{pred}$ corresponding to the final token predicted by the VLM backbone is used as the condition to guide the action model in converting this representation into robot motion trajectory $\tau$, which can be formulated as:
\begin{equation}
    \hat{\mathbf{\tau}}_t = \mathcal{A}_\theta\left(\textbf{h}_t^{pred} \right)
\end{equation}
where $\hat{\mathbf{\tau}}_t$ is the predicted motion trajectory in current timestamp $t$. The pseudocode presented in Alg.~\ref{alg:vla_video_cot} illustrates the complete online inference process of \ours in detail.

\begin{algorithm}[H]
\caption{\ours Online Inference}
\label{alg:vla_video_cot}
\begin{algorithmic}[1]
\State \textbf{Input:} Observation video stream $\mathcal{O} = \{\mathbf{o}_1, \mathbf{o}_2, \dots, \mathbf{o}_t\}$, Instruction $I$
\State \textbf{Initialize:} Memory $\mathcal{M} \gets \emptyset$, Visual Cache $\mathcal{V} \gets \emptyset$
\Procedure{OnlineInference}{$I, \mathcal{O}$}
    \While{\textbf{true}}
        \State $\mathbf{E}^I \gets \text{Tokenizer}(I)$
        \State $\mathbf{v}_t \gets \text{VisionEncoder}(\mathbf{o}_t)$ \Comment{Encode the current visual frame}
        \State $\mathcal{V} \gets \text{Cache}(\mathcal{V}, \mathbf{v}_t)$ \Comment{Update visual cache with the new feature}
        \State $\mathbf{E}^V \gets \text{Sampling\&Pooling}(\mathcal{V})$ \Comment{Obtain visual tokens from cache}
        \State $\mathbf{E}^T \gets \text{RoPE}(\Delta t)$ \Comment{Create temporal-aware indicator token}
        \State $\mathbf{E}^M \gets \text{Tokenizer}(\mathcal{M})$
        \State $\mathbf{E}^{\text{CoT}} \gets \text{LLM.forward}(\mathbf{E}^I, \mathbf{E}^T, \mathbf{E}^V, \mathbf{E}^M)$
        \If{$\mathbf{E}^{\text{CoT}} = \texttt{<think\_on>}$}
            \State $c_t \gets \text{LLM.generate}(\mathbf{E}^I, \mathbf{E}^T, \mathbf{E}^V, \mathbf{E}^M, \mathbf{E}^{\text{CoT}})$
            \State $\mathcal{M} \gets \text{UpdateMemory}(\mathcal{M}, c_t)$
        \EndIf
        \State $h_t^{\text{pred}} \gets \mathbf{E}^{\text{CoT}}[-1]$ \Comment{Use the hidden state of the last token as input for the action model}
        \State $\hat{\tau}_t \gets A_{\theta}(h_t^{\text{pred}})$ \Comment{Generate the next trajectory}
        
        \If{$\hat{\tau}_t = \texttt{stop}$}
            \State \textbf{break}
        \Else
            \State $\text{ExecuteAction}(\hat{\tau}_t)$
        \EndIf
    \EndWhile
\EndProcedure

\end{algorithmic}
\end{algorithm}

\section{Data Collection}
\label{sec:data}
\newcommand{\cmark}{\scalebox{1}{$\checkmark$}}
\newcommand{\xmark}{\scalebox{1}{$\times$}}

\begin{table*}[t]
    \centering
    \small
    \caption{Comparison of existing navigation datasets. Nav-AdaCoT-2.9M is the first dataset to integrate three navigation tasks (ObjNav, Track, ImageNav) and provide adaptive chain-of-thought reasoning.}
    \label{tab:datasets_comp}
    \setlength{\tabcolsep}{3pt}
    \begin{tabular}{l cc c cccc c c c}
        \toprule
        \multirow{2}{*}{Dataset} & \multicolumn{3}{c}{Scenes} & \multicolumn{4}{c}{Instruction Capability} & \multirow{2}{*}{$N_{step}$} & \multirow{2}{*}{$N_{cot}$} & \multirow{2}{*}{Action} \\
        \cmidrule(lr){2-4} \cmidrule(lr){5-8}
        & HM3D & MP3D & $N_{scene}$ & ObjNav & Track & ImageNav & Modality & & & \\
        \midrule
        HM3D ObjNav~\citep{ramakrishnan2021habitat} & \cmark & \xmark & 80 & \cmark & \xmark & \xmark & L & - & - & Des. \\
        MP3D ObjNav~\citep{chang2017matterport3d} & \xmark & \cmark & 56 & \cmark & \xmark & \xmark & L & - & - & Des. \\
        SOON~\citep{zhu2021soon} & \cmark & \xmark & 90 & \cmark & \xmark & \xmark & L & 30K & - & Des. \\
        HM3D OVON~\citep{yokoyama2024hm3d} & \cmark & \xmark & 181 & \cmark & \xmark & \xmark & L & 53K & - & Des. \\
        EVT-Bench~\citep{wang2025trackvla} & \cmark & \cmark & 703 & \xmark & \cmark & \xmark & L & 855K & - & Traj. \\
        HM3D ImgNav~\citep{krantz2023navigating} & \cmark & \xmark & 145 & \xmark & \xmark & \cmark & L & - & - & Des. \\
        OctoNav-Bench~\citep{gao2025octonav} & \cmark & \cmark & 438 & \cmark & \xmark & \cmark & V, L & 45K & 10K & Des. \\
        Nav-CoT-110K~\citep{liu2025nav} & \cmark & \cmark & 342 & \cmark & \xmark & \xmark & V, L & 110K & 110K & Des. \\
        \rowcolor{gray!20}
        \textbf{Nav-AdaCoT-2.9M (Ours)} & \cmark & \cmark & \textbf{718} & \cmark & \cmark & \cmark & V, L & \textbf{2.9M} & \textbf{472K} & Traj. \\
        \bottomrule
    \end{tabular}
\end{table*}

\begin{figure}[!t]
    \centering
    \includegraphics[width=0.7\linewidth]{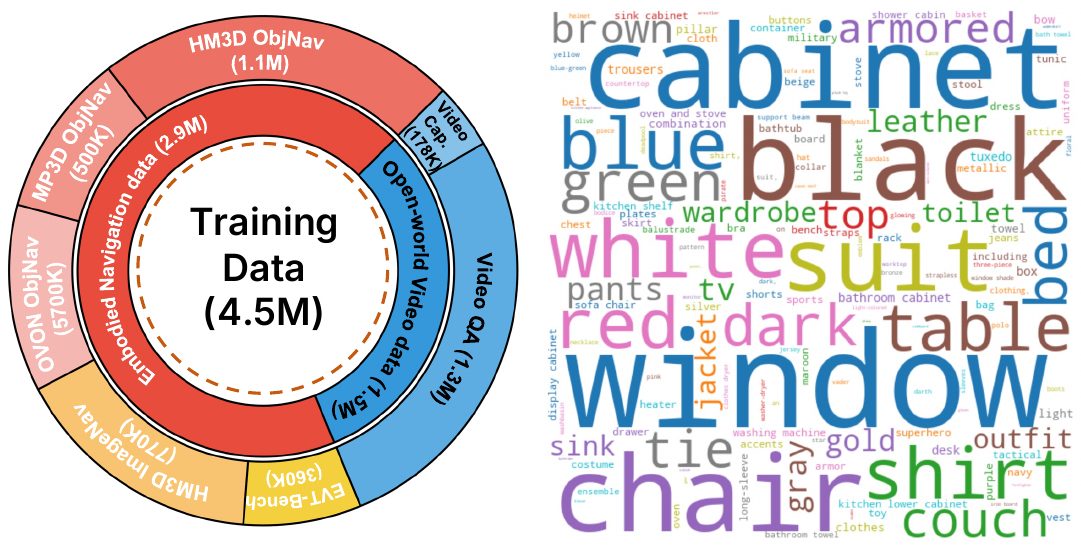}
    \caption{Data distribution and instruction word cloud for the \ours training dataset.}
    \label{fig:dataset}
\end{figure}

Our framework is trained on Nav-AdaCoT-2.9M, a large-scale dataset we constructed, supplemented by public open-world video datasets. Tab.~\ref{tab:datasets_comp} presents a statistical comparison between Nav-AdaCoT-2.9M and existing public embodied navigation datasets, evaluating metrics such as scene number, task type, total steps, CoT annotation number, and action modalities.
Notably, our dataset surpasses others in scene number, task variety, and input modality richness. It also features the largest count of CoT annotations to date. Furthermore, Nav-AdaCoT-2.9M employs trajectory-based annotation, which provides finer-grained supervision compared to discrete action-based datasets.

\subsection{Embodied Navigation Data}
\subsubsection{Navigation Data Generation}
To ensure both diversity and comparability, we construct our training data from several widely used embodied navigation benchmarks.

\textbf{Object-Goal Navigation.} We use data from three benchmarks:
\begin{itemize}
    \item HM3D ObjNav~\citep{ramakrishnan2021habitat}: For this category-level search task, we utilize a subset of the human demonstration data provided by Habitat-Web~\citep{ramrakhya2022habitat}.
    \item MP3D ObjNav~\citep{chang2017matterport3d}: We collect shortest-path trajectories to serve as training data.
    \item HM3D OVON~\citep{yokoyama2024hm3d}: For this zero-shot, open-vocabulary task, we also collect shortest-path trajectories.
\end{itemize}

\textbf{Visual Tracking.} We leverage EVT-Bench~\citep{wang2025trackvla} to curate a multi-person indoor tracking dataset.

\textbf{Image-Goal Navigation.} We use the HM3D Instance ImageNav~\citep{krantz2022instance} benchmark. For this, we also generate shortest-path trajectories and derive step-by-step action labels.

Leveraging these existing resources, we propose Nav-AdaCoT-2.9M, a large-scale dataset encompassing 2.9M step-by-step adaptive Chain-of-Thought trajectories. Distinct from prior datasets that predominantly furnish only instructions and expert action labels, Nav-AdaCoT-2.9M explicitly integrates structured reasoning that is aligned with observations and instructions. This design effectively bridges the domains of perception, language, and action. As the cornerstone for the supervised fine-tuning phase of \ours, this dataset facilitates the acquisition of structured reasoning capabilities in \ours prior to the reinforcement learning-based post-training.

\begin{figure*}[!t]
    \centering
    \includegraphics[width=1\linewidth]{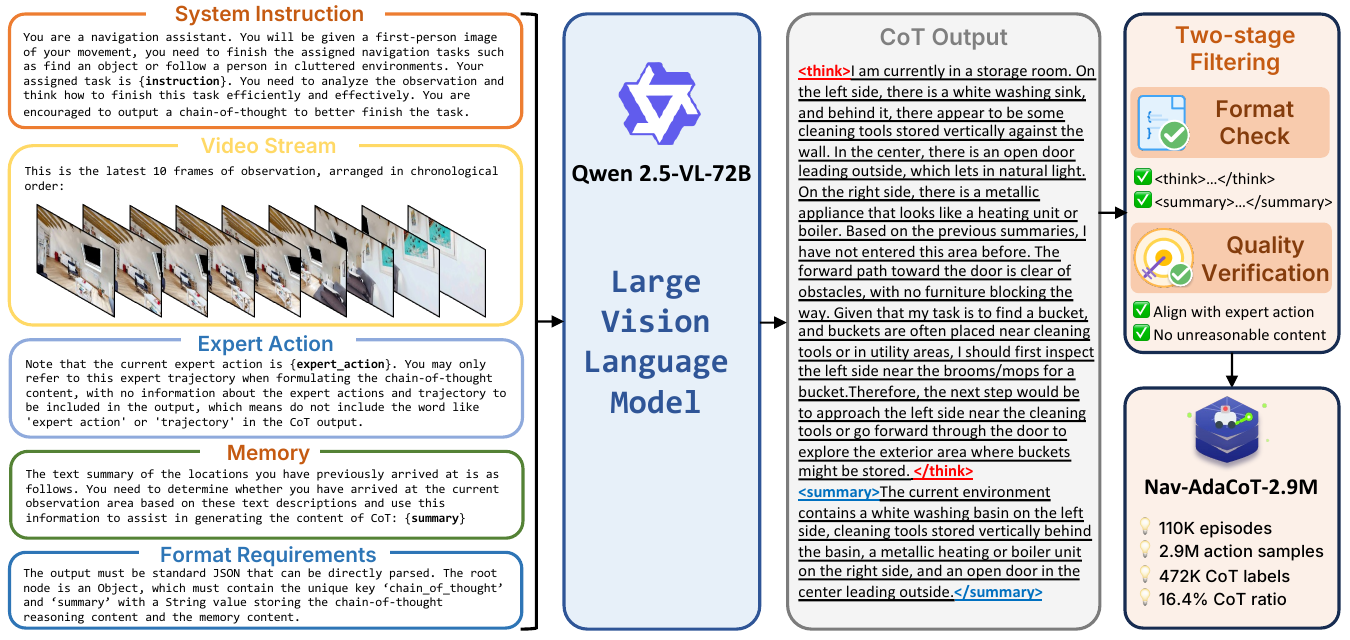}
    \caption{The autonomous adaptive CoT labeling pipeline of \ours.}
    \label{fig:cot}
\end{figure*}

\subsubsection{Autonomous Adaptive CoT labeling pipeline}
We propose an autonomous adaptive Chain-of-Thought data labeling pipeline, specifically designed to construct high-quality CoT labels for embodied navigation and reasoning tasks. This pipeline leverages the reasoning capabilities of Vision-Language Models to generate coherent, step-by-step CoT rationales that justify navigation decisions in complex environments. As illustrated in Fig.~\ref{fig:cot}, we applied adaptive CoT labeling to the entire embodied navigation dataset described in the previous section.

To generate high-quality Chain-of-Thought labels, we designed a composite prompt for Qwen2.5-VL-72B~\citep{bai2025qwen2} that incorporates five essential components: 1) navigation instructions, 2) egocentric visual stream input (the most recent 10 frames, included to reduce the computational load of the VLM), 3) prior memory content, 4) expert trajectories at each step, and 5) explicit formatting requirements. This prompt guides the VLM to reason about spatial relationships, environmental constraints, and the semantic meaning of instructions, thereby generating structured, step-by-step CoT sequences. The outputs follow a standardized format: reasoning processes are enclosed within $\texttt{<think>}\cdots\texttt{</think>}$ tags, while summaries are contained in $\texttt{<summary>}\cdots\texttt{</summary>}$ tags. This formatting ensures transparent alignment among observations, reasoning, and memory.
When this pipeline was executed across our diverse set of environments, approximately 472K CoT responses were generated from 2.9M samples. Each response includes a detailed CoT analysis and decision-making process for the navigation scenario, as well as linguistic memory describing the current environmental context. These raw outputs were further refined through a two-stage filtering procedure:
1) Rule-based checks: Responses that were incomplete or logically inconsistent were discarded.
2) Quality verification: Decisions were cross-validated against expert navigation trajectories to ensure accuracy.
Following refinement, we constructed the Nav-AdaCoT-2.9M dataset. Serving as the Supervised Fine-Tuning data for \ours, this dataset provides rich reasoning trajectories that tightly integrate perception, instruction following, and navigation decision-making.

\subsection{Open-World Video Data}
Furthermore, co-training with open-world video data has been shown in multiple studies~\citep{zhang2024uni,wei2025streamvln,black2025pi_} to enhance model generalization and reduce the sim-to-real transfer gap. Consistent with these findings, we incorporate a variety of publicly available open-world video datasets~\citep{azuma2022scanqa, zhang2024video, feng2025video} into our training data. Beyond prior efforts, our approach not only improves general visual understanding but also further strengthens adaptive reasoning through additional adaptive CoT annotations. Specifically, we utilize three datasets, LLaVA-Video-178K~\citep{zhang2024video}, Video-R1~\citep{feng2025video}, and ScanQA~\citep{azuma2022scanqa}, comprising a total of 1.6M samples, and construct an adaptive CoT-based video dataset by categorizing samples according to difficulty. In particular, the Video-R1 dataset, which contains relatively challenging video QA pairs, is organized as a CoT-annotated subset, whereas the other two datasets are formatted as non-CoT subsets. This design enables the model to further develop the ability to autonomously decide whether reasoning is required for a given input.

\subsection{Dataset Statistics}
Ultimately, the training dataset for \ours comprises the two aforementioned types of datasets, totaling 4.5M training samples. Specifically, it includes 2.9M samples of embodied navigation data and 1.6M samples of open-world video data, with the detailed data distribution illustrated in Fig.~\ref{fig:dataset}.
\section{Training Recipe}
\begin{figure}[!t]
    \centering
    \includegraphics[width=0.7\linewidth]{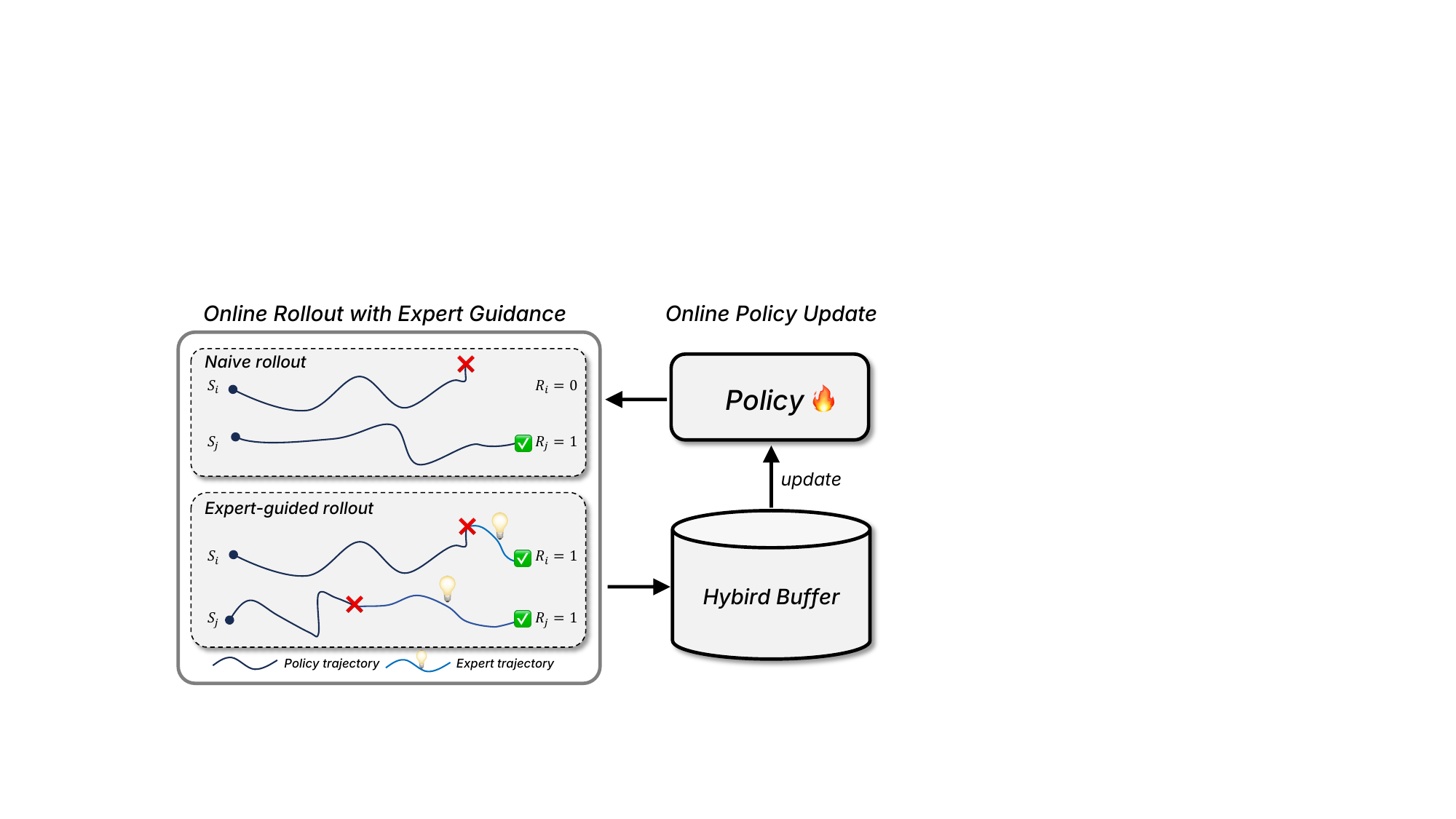}
    \caption{Online post-training with a hybrid rollout procedure.}
    \label{fig:rollout}
\end{figure}

\subsection{Model Pre-train}
The VLM backbone used in \ours does not natively support adaptive reasoning. To address this, we first conduct a pre-training stage on our custom open-world adaptive CoT video dataset (detailed in Sec.~\ref{sec:data}). Following standard VLM training paradigms, we fine-tune the model for a single epoch. This process equips the model with the foundational ability to perform adaptive visual reasoning. The training is supervised using a standard cross-entropy (CE) loss, applied at the token level.

\subsection{Supervised Fine-Tuning}
Following the pre-training phase, we perform supervised fine-tuning (SFT) to establish robust navigation and video reasoning capabilities. Specifically, we train the model using standard imitation learning on a combined dataset that integrates our embodied navigation data with the open-world video data. This co-training strategy ensures the model retains general-purpose visual reasoning while acquiring task-specific navigation skills. The training objective can be formalized as:
\begin{equation}\label{eq:sft}
\min_{\theta}\;\; \mathcal{L}_{\text{SFT}}(\theta)=\alpha\mathcal{L}_{\text{MSE}}(\hat{\tau}_t,\tau_t^{gt})+(1-\alpha)\mathcal{L}_{\text{CE}}(E_t^{pred},E_t^{gt})
\end{equation}
where $\mathcal{L}_{\text{MSE}}$ is the Mean Squared Error loss that supervises the predicted action trajectory $\hat{\tau}_t$ against the ground-truth trajectory $\tau^{gt}_t$, $\mathcal{L}_{\text{CE}}$ is the Cross-Entropy loss that supervises the generation of all textual outputs, including both the CoT reasoning and the VQA responses. $\alpha$ is a hyperparameter that balances the contribution of the two loss components.

\subsection{Online Expert-guided Post-training}
To address the limitations of offline imitation learning, such as covariate shift, and to better align the VLM's high-level representations with the closed-loop robot continuous action, we introduce an online post-training stage.
Starting from the SFT checkpoint, the agent actively interacts with the simulation environment to collect fresh, on-policy trajectories. The policy is then updated using a hybrid objective function. This objective combines outcome-driven optimization with expert-guided supervision. This dual approach allows the model to explore more effective strategies while preventing catastrophic forgetting of the expert policy.

\subsubsection{Probabilistic Continuous Action Model}
Existing VLA architectures often employ discrete tokenization for actions, which sacrifices precision; others use generative models like diffusion or flow matching, which incur high computational costs due to iterative denoising. To address this trade-off between high-precision continuous control and efficient inference, we propose a lightweight probabilistic projection head.

Let $\mathbf{h}_t$ denote the visual-linguistic features extracted from the VLM backbone at timestep $t$. We parameterize the policy $\pi_{\theta}(\mathbf{a}_t | \mathbf{s}_t)$ as a multivariate Gaussian distribution. Specifically, the action head projects $\mathbf{h}_t$ to predict the mean $\boldsymbol{\mu}_{\theta}(\mathbf{h}_t)$ and the logarithm of the standard deviation $\log \boldsymbol{\sigma}_{\theta}(\mathbf{h}_t)$:
\begin{equation}
\pi_{\theta}(\mathbf{a}_t | \mathbf{s}_t) = \mathcal{N}\left(\boldsymbol{\mu}_{\theta}(\mathbf{h}_t), \text{diag}\left(\boldsymbol{\sigma}_{\theta}(\mathbf{h}_t)^2\right)\right)
\end{equation}

During online post-training rollout, stochastic exploration is implemented by sampling actions $\mathbf{a}_t$ from the policy distribution $\mathbf{a}_t \sim \pi_\theta(\cdot \mid \mathbf{s}_t)$. In contrast, for the validation phase, deterministic execution is adopted, where actions $\mathbf{a}_t$ are set to the mean value $\mathbf{a}_t=\boldsymbol{\mu}_\theta(\mathbf{h}_t)$ of the policy’s action distribution conditioned on the hidden state $\mathbf{h}_t$.

\subsubsection{Hybrid Rollout}
To balance exploration with successful task completion, we employ a hybrid data collection strategy. As illustrated in Fig.~\ref{fig:rollout}, we alternate between two rollout modes:

\begin{itemize}
    \item Naive rollout: The current policy $\pi_{\theta}$ interacts with the environment independently. We store the complete interaction trajectories $\tau = \{(\mathbf{s}_t, \mathbf{a}_t, r_t)\}$, and only successful trajectories are filtered out and incorporated into the hybrid buffer. As on-policy data, this dataset accurately reflects the current policy's capabilities and provides high-quality positive examples for reinforcing successful action sequences.
    
    \item Expert-guided rollout: To address inefficient exploration and mitigate erroneous behaviors, the system incorporates an expert policy $\pi^*$ (implemented via a Shortest Path planner in simulator). When the agent triggers an irrational condition (\eg, oscillating or stuck for $k$ steps, here $k=15$) or eventually fails. By taking control and demonstrating a recovery path, the expert provides high-quality, corrective trajectories. These demonstrations are then added to the hybrid buffer, enriching it with valuable examples of how to escape difficult states and improving overall agent robustness.
\end{itemize}

\subsubsection{Online Fine-tuning with Augmented Loss}
Pure reinforcement learning can be unstable and sample-inefficient under sparse rewards and long horizons, while pure imitation learning may overfit to the expert state distribution and suffer from covariate shift. We therefore adopt a demonstration-augmented online post-training scheme~\citep{rajeswaran2017learning}, where interaction data provides an outcome-driven learning signal and expert-guided trajectories provide a stabilizing supervised signal. Specially, we optimize the following composite objective:
\begin{equation}
\label{eq:posttrain_obj}
\begin{aligned}
\min_{\theta}\;\; \mathcal{L}_{\text{post}}(\theta)
&= \lambda\mathcal{L}_{\text{RL}}(\theta) + (1-\lambda)\mathcal{L}_{\text{SFT}}(\theta) \\
\text{where }\;\; \mathcal{L}_{\text{RL}}(\theta)
&= -\mathbb{E}_t
\left[
\min\Big(
r_t(\theta) A_t,\;
\mathrm{clip}(r_t(\theta), 1-\epsilon, 1+\epsilon)\,A_t
\Big)
\right],
\end{aligned}
\end{equation}
where $\mathcal{L}_{\text{RL}}$ is a PPO-style policy-gradient objective，and we use REINFORCE++~\citep{hu2025reinforce++} to calculate the advantage $A_t$. $\mathcal{L}_{\text{SFT}}$ is the imitation loss defined in Eq.~\ref{eq:sft}.
\newpage
\subsection{Implementation Details}
\subsubsection{Training Details} 
\ours is trained on a cluster with 128 NVIDIA A100 GPUs using a three-stage training pipeline. In the first stage, we leverage open-world video data for pre-training to endow the model with adaptive general visual reasoning capabilities. Consistent with standard VLM practices~\citep{liu2023llava}, this pre-training runs for a single epoch. In the second stage, all embodied navigation data and open-world video data are mixed and randomly shuffled for co-training 20K steps with a total batch size of 512. In the online post-training phase, the policy undergoes 10 rollout iterations of updates using the training datasets of HM3D OVON, HM3D Instance ImageNav and EVT-Bench DT benchmarks. For each iteration, we use the current policy to collect 128 episodes of on-policy data, which are then added to the hybrid buffer before the model is updated. For open-world video data, all videos are sampled at 1 FPS to reduce redundancy between consecutive frames. Throughout all stage training, only the visual encoder’s parameters are frozen; all other components are updated. The hyperparameters are set to $\alpha=0.5$ and $\lambda=0.01$, which is determined by the scale of different losses.

\subsubsection{Inference Details} 
During inference, we maintain a compact and consistent model architecture by not using task-specific tokens for task partitioning. Instead, at each step, the model autoregressively predicts a CoT indicator token. Based on this indicator, it may then generate CoT content. Finally, the hidden state corresponding to the last generated token is fed into the action module, which predicts the robot's future motion trajectory.
\section{Experiments}
To comprehensively evaluate the performance of \ours, we conducted a series of extensive experiments in both simulation and the real world. We first quantitatively compare \ours against state-of-the-art methods on several standard embodied navigation benchmarks. We then conduct detailed ablation studies to validate the effectiveness of each of our proposed key components. Furthermore, we have validated that the proposed model framework and training recipe in \ours demonstrate emergent generalization capabilities across diverse domains and tasks. Finally, we demonstrate \ours's ability to transfer to a real-world robot and complete practical navigation tasks in a zero-shot manner, verifying its real-world generalization and utility.

\subsection{Experiment Setups}
\subsubsection{Benchmarks.}
Our method is evaluated on multiple public benchmarks, including those for Object Goal Navigation (HM3Dv1 ObjNav, HM3Dv2 ObjNav, MP3D ObjNav, and HM3D OVON), Embodied Visual Tracking (EVT-Bench), and Image Goal Navigation (HM3D Instance ImageNav). Notably, a shared model checkpoint is used across all tasks, with no additional fine-tuning performed for any individual task.

\subsubsection{Baselines.}
We conduct a comprehensive comparison of \ours against current state-of-the-art models, categorized into three groups: (1) modular methods, often separate the model into perception, mapping and planning, \ie~\citep{yokoyama2024hm3d,yin2025unigoal,yin2024sg,yu2023l3mvn,zhang2025embodied,chang2023goat,krantz2023navigating,zhang2025apexnav}, (2) end-to-end small-scale models often leverage a pre-trained network for visual feature extraction, which are then integrated with a policy network to output robot actions, \ie~\citep{ramrakhya2022habitat, ramrakhya2023pirlnav,yokoyama2024hm3d,yokoyama2025film, zhong2024empowering, zeng2024poliformer}, and (3) VLA models~\citep{zhang2024uni,wang2025trackvla,zhang2025embodied,liu2025trackvla++,liu2025nav,zhu2025mtu}.

\subsubsection{Metrics.}
To evaluate navigation performance, we use standard metrics from public benchmarks, including Success Rate (SR), Success-weighted Path Length (SPL), Tracking Rate (TR), and Collision Rate (CR). 

\newpage
\subsection{Simulation Experiments}
\subsubsection{Object Goal Navigation}
\begin{table}[!t]
\centering
\small
\caption{\textbf{Performance on object goal navigation.} Comparison on HM3D ObjNav~\citep{ramakrishnan2021habitat} and MP3D ObjNav~\citep{chang2017matterport3d} benchmarks. The \textbf{best} and the \underline{second best} results are denoted by \textbf{bold} and \underline{underline}.}
\begin{tabular}{lcccccc}
\toprule
\multirow{2}{*}{Method} & \multicolumn{2}{c}{\textbf{\textit{HM3Dv1}}} & \multicolumn{2}{c}{\textbf{\textit{HM3Dv2}}} & \multicolumn{2}{c}{\textbf{\textit{MP3D}}} \\
\cmidrule(lr){2-3} \cmidrule(lr){4-5} \cmidrule(lr){6-7}
& SR$\uparrow$ & SPL$\uparrow$ & SR$\uparrow$ & SPL$\uparrow$ & SR$\uparrow$ & SPL$\uparrow$ \\ 
\midrule
VLFM~\citep{yokoyama2024vlfm} & 52.5 & 30.4 & 63.6 & 32.5 & 36.4 & 17.5 \\
SG-Nav~\citep{yin2024sg} & 54.0 & 24.9 & 49.6 & 25.5 & 40.2 & 16.0  \\
L3MVN~\citep{yu2023l3mvn} & 54.2 & 25.5 & 36.6 & 15.7 & - & - \\
UniGoal~\citep{yin2025unigoal} & 54.5 & 25.1 & - & - & 41.0 & 16.4  \\
Habitat-Web~\citep{yokoyama2024hm3d} & 57.6 & 23.8 & 31.6 & 8.5 \\
InstructNav~\citep{long2024instructnav} & - & - & 58.0 & 20.9 & - & - \\
ApexNav~\citep{zhang2025apexnav} & 59.6 & 33.0 & 76.2 & 38.0 & 39.2 & \underline{17.8} \\
OVRL~\citep{yadav2023offline} & 62.0 & 26.8 & - & - & 28.6 & 7.4 \\
OVRL-v2~\citep{yadav2023ovrl} & 62.8 & 28.1 & - & - & - & - \\
LFG~\citep{shah2023navigation} & 68.9 & 36.0 & - & - & - & -\\
PirlNav~\citep{ramrakhya2023pirlnav} & 70.4 & 34.1 & - & - & - & - \\
FiLM-Nav~\citep{yokoyama2025film} & 61.7 & 37.3 & \underline{77.0} & \textbf{41.3} & - & - \\
CogNav~\citep{cao2024cognav} & 72.5 & 26.2 & - & - & 46.6 & 16.1 \\
Uni-NaVid~\citep{zhang2024uni} & \underline{73.7} & 37.1 & - & - & - & - \\
\midrule
\rowcolor{gray!20}
\textbf{\ours(SFT)} & 70.6 & \underline{38.2} & 76.4 & 32.6 & \underline{47.4} & \underline{25.8} \\
\rowcolor{gray!20}
\textbf{\ours} & \textbf{79.1} & \textbf{42.9} & \textbf{83.0} & \underline{40.5} & \textbf{58.9} & \textbf{26.5} \\
\bottomrule
\end{tabular}

\label{tab:objnav_hm3d}
\end{table} 

\begin{table}[!t]
\centering
\small
\caption{\textbf{Performance on object goal navigation.} Comparison on HM3D-OVON~\citep{yokoyama2024hm3d} benchmark. The \textbf{best} and the \underline{second best} results are denoted by \textbf{bold} and \underline{underline}.}
\begin{tabular}{lcccccc}
\toprule
\multirow{2}{*}{Method} & \multicolumn{2}{c}{\textbf{\textit{Val Seen}}}   & \multicolumn{2}{c}{\makecell{\textbf{\textit{Val Seen}} \\ \textbf{\textit{Synonyms}}}} & \multicolumn{2}{c}{\textbf{\textit{Val Unseen}}} \\
\cmidrule(lr){2-3} \cmidrule(lr){4-5} \cmidrule(lr){6-7}
& SR$\uparrow$ & SPL$\uparrow$ & SR$\uparrow$ & SPL$\uparrow$ & SR$\uparrow$ & SPL$\uparrow$ \\
\midrule
BC & 11.1 & 4.5 & 9.9 & 3.8 & 5.4 & 1.9 \\
DAgger & 11.1 & 4.5 & 9.9 & 3.8 & 5.4 & 1.9 \\
RL & 18.1 & 9.4 & 15.0 & 7.4 & 10.2 & 4.7 \\
BCRL & 39.2 & 18.7 & 27.8 & 11.7 & 18.6 & 7.5 \\
DAgRL & 41.3 & 21.2 & 29.4 & 14.4 & 18.3 & 7.9 \\
VLFM~\citep{yokoyama2024vlfm} & 35.2 & 18.6 & 32.4& 17.3 & 35.2 & 19.6 \\
DAgRL$+$OD~\citep{yokoyama2024hm3d} & 38.5 & 21.1 & 39.0 & 21.4 & 37.1 & 19.8 \\
Uni-NaVid~\citep{zhang2024uni} & 41.3 & 21.1 & 43.9 & 21.8 & 39.5 & 19.8 \\
TANGO~\citep{ziliotto2025tango} & - & - & - & - & 35.5 & 19.5 \\
FiLM-Nav~\citep{yokoyama2025film} & 44.9 & 24.5 & 40.1 & 23.1 & 40.8 & 24.4 \\
MTU3D~\citep{zhu2025mtu} & 55.0 & 23.6 & 45.0 & 14.7 & 40.8 & 12.1 \\
NavFoM~\citep{zhang2025embodied} &  37.7 & 25.5 & 43.3 & \underline{29.9} & \underline{43.6} & \textbf{31.3} \\ 
Nav-R1~\citep{liu2025nav} & \underline{58.4} & 26.3 & \underline{48.1} & 23.1 & 42.2 & 20.1 \\
\midrule
\rowcolor{gray!20}
\textbf{\ours(SFT)} & 45.9 & \underline{26.5} & 44.8 & 27.1 & 41.5 & 22.4 \\
\rowcolor{gray!20}
\textbf{\ours} & \textbf{59.3} & \textbf{29.7} & \textbf{56.8} & \textbf{30.1} & \textbf{50.1} & \underline{24.6} \\
\bottomrule
\end{tabular}

\label{tab:objnav-ovon}
\end{table} 

First, we compared the performance metrics of \ours with those of state-of-the-art methods on the Object Goal Navigation task. Specifically, our evaluations were conducted across multiple publicly available benchmarks including HM3Dv1, HM3Dv2, MP3D and HM3D~OVON. 

As presented in Tab.~\ref{tab:objnav_hm3d}, \ours achieved SOTA performance on three closed-vocabulary benchmarks, significantly outperforming prior methods on both SR and SPL metrics. On HM3Dv1, \ours reaches 79.1 SR and 42.9 SPL, improving over previous SOTA video-based VLA model Uni-NaVid (73.7/37.1) by +5.4 SR (+7.3\%) and +3.9 SPL (+15.6\%). A comparable performance improvement is also observed in HM3Dv2, where \ours achieves 83.0 SR and 40.5 SPL, surpassing FiLM-Nav (77.0/41.3) by +6.0 SR (+7.8\%). It is noted that the SPL result achieved by our method is slightly lower than that of FiLM-Nav. This discrepancy primarily arises because the FiLM-Nav model only selects the next frontier position and then relies on a shortest path planner to reach it—an approach that confers greater advantages in the simulator compared to our method, which directly outputs trajectory-based actions. On the MP3D benchmark—where long-range exploration scenarios predominate—\ours achieves an SR of 58.9 and an SPL of 26.5. These results significantly outperform those of the prior SOTA methods CogNav (46.6/16.1) and ApexNav (39.2/17.8). Specifically, our method yields an +26.4\% improvement in SR and a substantial +32.8\% enhancement in SPL. This impressive result demonstrates that \ours possesses robust exploration and memory capabilities, validating its effectiveness in complex long-range navigation tasks. Collectively, these results show that \ours not only exhibits enhanced object-exploration capabilities in diverse and challenging unseen environments, but also produces substantially shorter and more efficient trajectories across benchmark tests, highlighting the benefits of adaptive reasoning and long-horizon linguistic memory in the Object Goal Navigation task.

To further validate the generalization capability of \ours, we evaluate its performance on HM3D OVON—a more challenging open-vocabulary object navigation benchmark. This benchmark comprises three distinct test splits: (1) \textit{val seen}, which includes object categories present in the training set; (2) \textit{val seen synonym}, which consists of goal categories synonymous with those encountered during training; and (3) \textit{val unseen}, which contains object categories not present in the training dataset. As illustrated in Tab.~\ref{tab:objnav-ovon}, \ours achieves the best performance across all three test splits, with SRs improved by 0.9 (+1.5\%), 8.7 (+18.1\%), and 6.6 (+15.1\%) respectively compared to the previous SOTA methods. This result demonstrates the strong cross-domain generalization capability of \ours.

\subsubsection{Embodied Visual Tracking}
\begin{table}[htbp]
    \centering
    \small
    \caption{\textbf{Performance on embodied visual tracking.} Comparison on EVT-Bench~\citep{wang2025trackvla}. $\dag$: Use GroundingDINO~\citep{liu2023grounding} as the open-vocabulary detector. $\ddag$: Use SoM~\citep{yang2023set} with GPT-4o~\citep{openai2024introducing} as the visual foundation model.}
    \begin{tabular}{lcccccc}
        \toprule
        \multirow{2}{*}{Method} & \multicolumn{3}{c}{\textbf{\textit{Single Target Tracking}}} & \multicolumn{3}{c}{\textbf{\textit{Distracted Tracking}}} \\
        \cmidrule(lr){2-4} \cmidrule(lr){5-7}
        & SR$\uparrow$ & TR$\uparrow$ & CR$\downarrow$ & SR$\uparrow$ & TR$\uparrow$ & CR$\downarrow$ \\
        \midrule
        IBVS$\dag$~\citep{gupta2016novel} & 42.9 & 56.2 & 3.75 & 10.6 & 28.4 & 6.14 \\
        PoliFormer$\dag$~\citep{zeng2024poliformer} & 4.67 & 15.5 & 40.1 & 2.62 & 13.2 & 44.5 \\
        EVT~\citep{zhong2024empowering} & 24.4 & 39.1 & 42.5 & 3.23 & 11.2 & 47.9 \\
        EVT$\ddag$~\citep{zhong2024empowering} & 32.5 & 49.9 & 40.5 & 15.7 & 35.7 & 53.3 \\
        Uni-NaVid~\citep{zhang2024uni} & 53.3 & 67.2 & 12.6 & 31.9 & 50.1 & 21.3 \\				
        TrackVLA~\citep{wang2025trackvla} & 85.1 & 78.6 & \underline{1.65} & 57.6 & 63.2 & 5.80 \\
        NavFoM~\citep{zhang2025embodied} & 86.0 & 80.5 & - & 61.4 & 68.2 & - \\
        NavFoM*~\citep{zhang2025embodied} & 88.4 & 80.7 & - & 62.0 & 67.9 & - \\
        TrackVLA++~\citep{liu2025trackvla++} & 86.0 & \underline{81.0} & 2.10 & \underline{66.5} & 68.8 & \textbf{4.71} \\
        \midrule
        \rowcolor{gray!20}
        \textbf{\ours(SFT)} & \underline{87.2} & 78.9 & \textbf{1.23} & 66.1 & \underline{69.7} & \underline{4.78} \\
        \rowcolor{gray!20}
        \textbf{\ours} & \textbf{88.4} & \textbf{81.2} & 2.07 & \textbf{67.6} & \textbf{73.5} & 5.51 \\
        \bottomrule
    \end{tabular}

    \label{tab:evt-bench}
\end{table}
    
To evaluate the efficacy of the proposed method for the Embodied Visual Tracking (EVT) task, we conduct a comprehensive comparative analysis on the EVT-Bench. Specifically, we evaluate two representative and challenging splits:
(1) \textit{Single-Target Tracking}: The agent must continuously track a single designated target in complex unseen environments.
(2) \textit{Distracted Tracking}: A more complex scenario in which the agent must sustain stable tracking of the correct target under instructions while resisting interference from multiple distractors.
As illustrated in Tab.~\ref{tab:evt-bench}, \ours demonstrates SOTA performance across both splits. In the \textit{Single Target Tracking} task, \ours achieves an SR of 88.4 and a TR of 81.2, matching or slightly surpassing the previous best methods like NavFoM and TrackVLA++.  In the more challenging \textit{Distracted Tracking} scenario, \ours establishes a clear advantage, achieving an SR of 67.6 and a TR of 73.5. This represents a significant improvement of 1.1 (+1.7\%) in SR and 4.7 (+6.8\%) in TR compared to the previous SOTA method TrackVLA++. Notably, \ours outperforms NavFoM with multi-view setting while using only a monocular camera, demonstrating robust tracking and precise recognition. These results strongly validate the superior tracking capability and robustness of our approach, especially in complex environments with distractors.

\begin{figure*}[!t]
    \centering
    \includegraphics[width=1\linewidth]{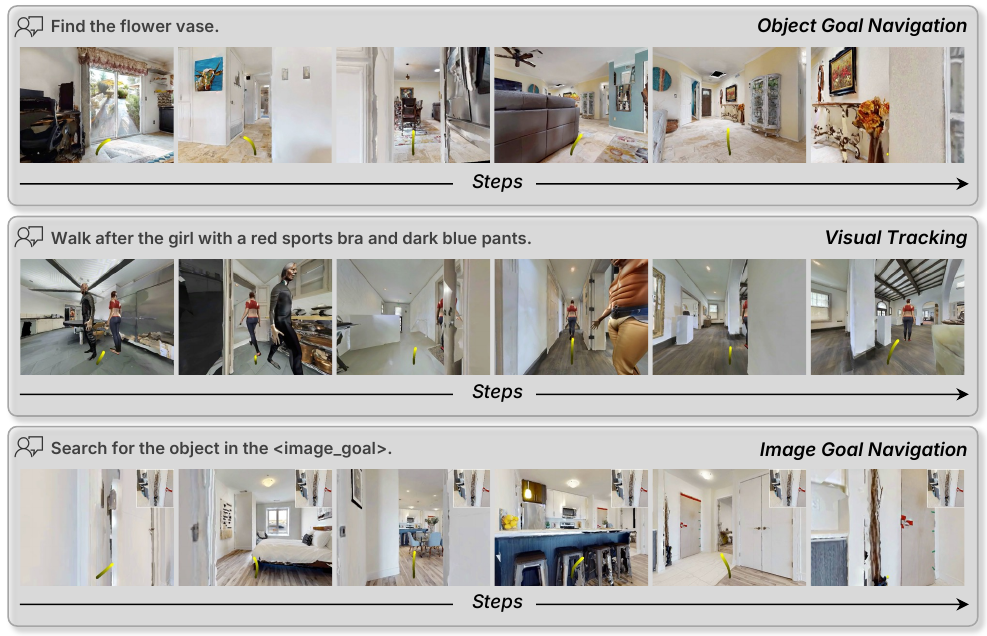}
    \caption{Performance visualization of \ours across various navigation benchmarks.}
    \label{fig:simulation_vis}
\end{figure*}

\subsubsection{Image Goal Navigation}
\begin{table}[!t]
    \centering
    \small
    \caption{\textbf{Performance on image goal navigation.} Comparison on HM3D Instance ImageNav benchmark~\citep{ramakrishnan2021habitat}. The \textbf{best} and the \underline{second best} results are denoted by \textbf{bold} and \underline{underline}.}
    \label{tab:imgnav}
    \begin{tabular}{lcc}
        \toprule
        Method & \multicolumn{2}{c}{\textbf{\textit{HM3D Instance ImageNav}}} \\
        \cmidrule(lr){2-3}
        & SR$\uparrow$ & SPL$\uparrow$ \\
        \midrule
        Krantz~\etal~\citep{krantz2022instance} & 8.3 & 3.5 \\
        OVRL-v2-IIN~\citep{yadav2023ovrl} & 24.8 & 11.8 \\
        PSL~\citep{sun2024prioritized} & 23.0 & 11.4 \\
        GOAT~\citep{chang2023goat} & 37.4 & 16.1 \\
        Mod-IIN~\citep{krantz2023navigating} & 56.1 & 23.3 \\
        UniGoal~\citep{yin2025unigoal} & \underline{60.2} & 23.7 \\
        \midrule
        \rowcolor{gray!20}
        \textbf{\ours(SFT)} & 51.1 & \underline{32.6} \\
        \rowcolor{gray!20}
        \textbf{\ours} & \textbf{60.8} & \textbf{37.4} \\
        \bottomrule
    \end{tabular}
\end{table}

To further evaluate the Image Goal Navigation capabilities of our model, we evaluate \ours on the HM3D Instance ImageNav benchmark. This task requires the agent to navigate to a specific object instance depicted in a goal image within complex and unseen environments. As presented in Table~\ref{tab:imgnav}, \ours achieves state-of-the-art results on this benchmark. It achieves an SR of 60.8, which is slightly higher than the previous SOTA method UniGoal (60.2/23.7). Note that, UniGoal leverages the LightGlue~\citep{lindenberger2023lightglue} keypoint-matching algorithm as an additional criterion, whereas \ours relies solely on the model’s implicit reasoning. More impressively, \ours demonstrates a substantial improvement in navigation efficiency, achieving an SPL of 37.4. This represents a remarkable 13.7 (+57.8\%) improvement over UniGoal. This significant gain in SPL underscores our model's ability to not only successfully find the target instance but also to do so via much more direct and efficient paths, highlighting the advanced reasoning and planning abilities of \ours.

\subsubsection{Visualization Results}
Fig.~\ref{fig:simulation_vis} illustrates several visualizations of \ours on the simulation benchmarks, including the robot’s egocentric visual observations, a top-down scene map, input instructions, and the outputs of adaptive CoT and the predicted trajectory. These examples show that our model efficiently accomplishes multiple embodied navigation tasks while adaptively generating CoT reasoning. This not only enhances interpretability but also improves the overall quality of the navigation process.

\subsection{Real-World Experiments}
\subsubsection{Robot Platform and Deployments}
\begin{figure}[!t]
    \centering
    \begin{minipage}[t]{0.4\linewidth}
        \centering
        \includegraphics[width=\linewidth]{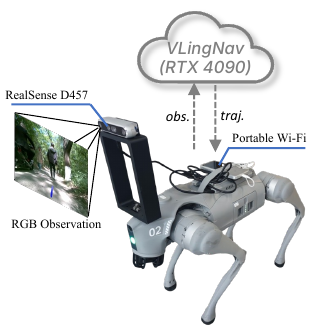}
        \caption{Real-world robot platform setup.}
        \label{fig:robot}
    \end{minipage}
    \hspace{10pt}
    \begin{minipage}[t]{0.4\linewidth}
        \centering
        \includegraphics[width=\linewidth]{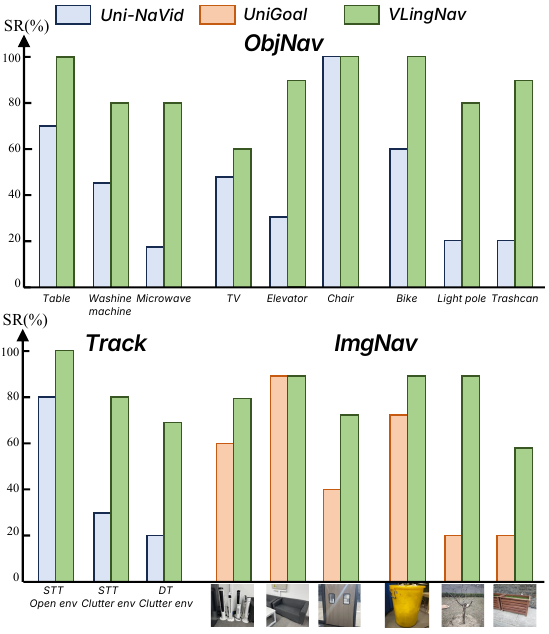}
        \caption{Real-world experiment results.}
        \label{fig:real_exp}
    \end{minipage}
\end{figure}

We provide a visualization of our robot platform in Fig.~\ref{fig:robot}. The platform is based on the Unitree Go2 quadruped robot, equipped with an Intel RealSense D457 camera on its head. In our work, we only utilize the RGB frames with a resolution of 1280$\times$800 from the camera, under a horizontal field of view (HFOV) of 90$^{\circ}$. Additionally, a portable Wi-Fi is mounted on the back of the robot to enable communication with the remote server through the Internet.

\ours is deployed on a remote server equipped with an NVIDIA RTX 4090 GPU. During real-world deployment, the server receives the instructions and images captured by the camera via the Internet. To ensure efficient communication, the images are compressed before transmission. After receiving the incoming data, the model performs inference and predicts the future trajectory, which is then transmitted to the quadruped robot for execution. Given that real-world navigation is an online process, we cache visual tokens from historically observed images. As a result, at each step the model only encodes the latest frame, which significantly improves inference efficiency. Furthermore, by leveraging \ours's visual memory compression strategy, our model maintains an inference latency of under 300~ms across 500 video frames. Including communication overhead (approximately 100~ms), \ours achieves an effective inference speed of around 2.5~FPS during long-horizon, real-world robot experiments.

Upon receiving the predicted trajectory, the robot employs a nonlinear model predictive control (NMPC) module for trajectory tracking~\citep{grandia2023perceptive}. Formulating the task as an optimization problem based on a kinematic unicycle model, the controller computes optimal linear and angular velocities over a receding horizon. 

\subsubsection{Object Goal Navigation}
We evaluated the Object Goal Navigation performance of \ours against the SOTA method Uni-NaVid, across three representative scenarios: home, office, and outdoor environment.
For each scenario, we selected three distinct target objects: (i) the table, washing machine, and microwave for the home environment; (ii) the TV, elevator, and trashbin for the office environment; (iii) the bike, light pole, and tree for the outdoor environment.
To mitigate the effects of randomness, we conducted 10 repeated trials for each target object. As shown in Fig.~\ref{fig:real_exp}, \ours achieves a significantly higher success rate than Uni-NaVid across all tested scenarios. These results validate the robust object recognition, exploration, and cross-scenario generalization capabilities of our model.

\subsubsection{Embodied Visual Tracking}
We evaluated the embodied visual tracking performance of our method against Uni-NaVid across three representative scenarios: (i) single-target tracking in open spaces, (ii) single-target tracking in cluttered indoor environments, and (iii) distracted tracking in crowded scenes with frequent occlusions and nearby distractors. To mitigate randomness, we conducted 10 repeated trials per scenario. As shown in Fig.~\ref{fig:real_exp}, our method consistently outperforms Uni-NaVid in tracking success rate, with the largest margins appearing in the distracted setting where transient occlusions and target switches are common. These results validate the effectiveness of our adaptive reasoning for re-identification after occlusion and the benefit of precise trajectory control, highlighting strong generalization to dynamic, cluttered environments.

\subsubsection{Image Goal Navigation}
We further evaluated Image Goal Navigation by comparing our method with UniGoal across three representative scene categories—home, office, and outdoor environments. For each category, we selected two image-specified goals and conducted 10 repeated trials per goal. As shown in Fig.~\ref{fig:real_exp}, our approach achieves a substantially higher success rate than UniGoal in all categories. These results suggest that multi-task training induces robust cross-modal grounding from text to images, while the combination of Adaptive CoT and linguistic memory supports reliable localization and efficient long-horizon navigation to visually specified targets under variations in camera intrinsics, viewpoints, and lighting.

\begin{figure*}[!t]
    \centering
    \includegraphics[width=0.96\linewidth]{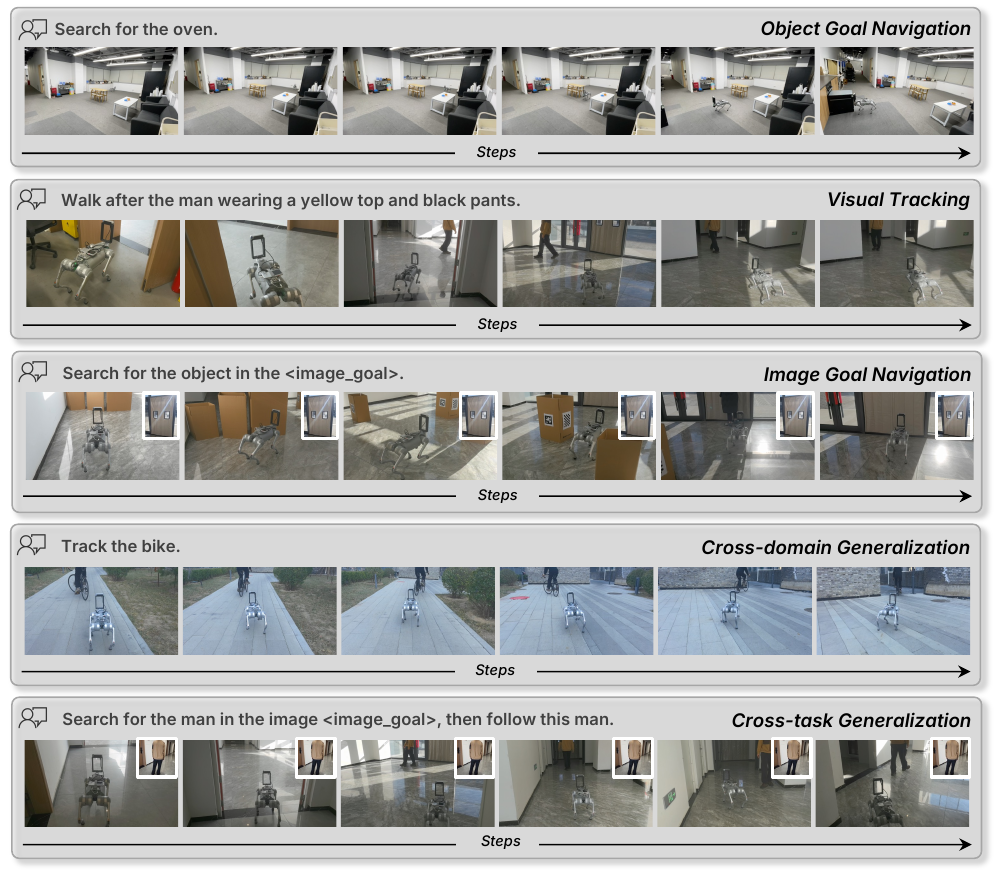}
    \caption{Qualitative performance of \ours in real-world deployments.}
    \label{fig:real_vis}
\end{figure*}

\subsubsection{Visualization Results}
Real-world experimental results are shown in Fig.~\ref{fig:real_vis}, where we evaluate the navigation capabilities of \ours under challenging scenarios. Specifically, we test the model across three representative scenario categories: office environments, household settings, and outdoor scenes. Within each category, we assess three core capabilities: object goal navigation, embodied visual tracking, and image goal navigation.
Notably, the model weights deployed on the real-world robot are the same as those used in the simulation experiments described in the previous section; no additional fine-tuning on real-world data is performed.
The results demonstrate that \ours exhibits strong sim-to-real transfer in both recognition and planning while sustaining high-frequency inference in real-world scenarios, enabling zero-shot deployment in complex environments.

\subsection{Emergence of Cross-Task and Cross-Domain Capabilities}
Joint training on multi-task navigation datasets leads \ours to exhibit emergent behaviors that generalize beyond any single task, yielding both cross-task and cross-domain capabilities.

\subsubsection{Cross-task Performance}
We observe clear cross-task transfer in real-world experiments (Fig.~\ref{fig:real_vis}). For instance, although the visual tracking task contains only language-format instructions, \ours can directly track targets specified by image goals in a zero-shot manner. Moreover, it composes behaviors across tasks: (1) Search for a language-described target, then switch to tracking that target. (2) Search for the target in the image goal and subsequently track it after locating it.
This compositionality arises from the VLA model’s shared, unified architecture and co-training on multi-task navigation datasets. Together, these factors enable the model to learn common navigation priors and transfer them successfully across diverse navigation tasks.

\subsubsection{Cross-domain Performance}
Second, we observe robust cross-domain generalization. Although trained only to track humans, \ours reliably tracks dynamic non-human targets. Moreover, \ours successfully localizes and navigates to out-of-distribution objectives specified only by fine-grained textual instructions, including category-ambiguous objects disambiguated by color, spatial constraints, or detailed descriptions. These behaviors indicate that multi-task learning, when co-trained with general visual understanding data, can substantially enhance \ours's generalization across domains.

\subsection{Ablation Studies}
To evaluate the contribution of each component and training strategy in \ours, we conducted comprehensive ablation studies. These studies were performed on the ObjectNav task using the HM3D OVON \textit{val unseen} benchmark, the EVT task using the EVT-Bench \textit{Distracted Tracking} benchmark, and the ImageNav task using the HM3D Instance ImageNav \textit{val} benchmark. For consistency, we adhered to the same training procedures and evaluation settings as those used for the full model. Below, we summarize the empirical results and analyze the key findings.

\subsubsection{\textbf{Adaptive CoT}}
\begin{table}[htbp]
    \centering
    \small
    \caption{Ablation study on Chain-of-Thought strategies. $r_{CoT}$ indicates the average percentage of steps where CoT reasoning is activated.}
    \label{tab:ablation_cot}
    
    \scalebox{1}{
        \setlength{\tabcolsep}{8pt}
        \begin{tabular}{l cc ccc cc c}
            \toprule
            \multirow{2}{*}{\textbf{CoT Strategy}} & \multicolumn{2}{c}{\textbf{ObjNav}} & \multicolumn{3}{c}{\textbf{Track}} & \multicolumn{2}{c}{\textbf{ImageNav}} & \multirow{2}{*}{$\boldsymbol{r_{CoT}}$ (\%)} \\
            \cmidrule(lr){2-3} \cmidrule(lr){4-6} \cmidrule(lr){7-8}
             & SR$\uparrow$ & SPL$\uparrow$ & SR$\uparrow$ & TR$\uparrow$ & CR$\downarrow$ & SR$\uparrow$ & SPL$\uparrow$ & \\
            \midrule
            
            w/o CoT & 36.2 & 16.5 & 62.7 & 68.5 & 6.28 & 56.3 & 27.3 & 0.0 \\
            Dense CoT (Per-step) & 25.3 & 13.0 & 59.8 & 70.1 & 26.3 & 19.6 & 13.2 & 100.0 \\
            Fixed Interval ($k=5$) & 42.5 & 23.5 & \textbf{68.5} & \textbf{74.2} & 9.18 & 48.2 & 28.7 & 20.0 \\
            Fixed Interval ($k=20$) & 39.7 & 19.4 & 66.2 & 70.8 & 11.9 & 51.3 & 31.2 & 5.0 \\
            
            \midrule
            \rowcolor{gray!20}
            \textbf{Adaptive CoT (Ours)} & \textbf{50.1} & \textbf{24.6} & 67.6 & 73.5 & \textbf{5.51} & \textbf{60.8} & \textbf{37.4} & \textbf{2.1} \\
            \bottomrule
        \end{tabular}
    }
\end{table}
To assess the impact of different reasoning strategies, we conducted an ablation study detailed in Tab.~\ref{tab:ablation_cot}. The results show that both a complete lack of reasoning (``w/o CoT'') and exhaustive reasoning at every step (``Dense CoT'') lead to suboptimal performance. While fixed-interval reasoning provides a moderate improvement, it remains inflexible. Our proposed Adaptive CoT strategy demonstrates clear superiority. It achieves the highest performance across all benchmarks. Remarkably, it accomplishes this while maintaining an exceptionally low reasoning frequency ($r_{CoT} = \text{2.1}\%$), far more efficient than even the sparse fixed-interval method. This highlights that dynamically and intelligently activating reasoning only when needed is crucial for creating high-performing, efficient embodied agents.

\subsubsection{\textbf{Visual-assisted Linguistic Memory}}
\begin{table}[htbp]
    \centering
    \small
    \caption{Ablation study on memory modalities.}
    \label{tab:ablation_memory}
    
    \scalebox{1}{
        \setlength{\tabcolsep}{10pt}
        \begin{tabular}{l cc ccc cc}
            \toprule
            \multirow{2}{*}{\textbf{Memory Mode}} & \multicolumn{2}{c}{\textbf{ObjNav}} & \multicolumn{3}{c}{\textbf{Track}} & \multicolumn{2}{c}{\textbf{ImageNav}} \\
            \cmidrule(lr){2-3} \cmidrule(lr){4-6} \cmidrule(lr){7-8}
             & SR$\uparrow$ & SPL$\uparrow$ & SR$\uparrow$ & TR$\uparrow$ & CR$\downarrow$ & SR$\uparrow$ & SPL$\uparrow$ \\
            \midrule
            
            w/o Memory & 15.4 & 3.5 & 37.5 & 59.1 & \textbf{1.90} & 21.0 & 3.7 \\
            Visual-only & 45.2 & 20.3 & 66.8 & 70.6 & 7.85 & 57.9 & 33.7 \\
            Language-only & 18.8 & 4.4 & 40.2 & 55.2 & 3.25 & 23.3 & 7.5 \\
            
            \midrule
            \rowcolor{gray!20}
            \textbf{VLingMem (Ours)} & \textbf{50.1} & \textbf{24.6} & \textbf{67.6} & \textbf{73.5} & 5.51 & \textbf{60.8} & \textbf{37.4} \\
            \bottomrule
        \end{tabular}
    }
\end{table}
We conducted an ablation study to evaluate the VLingMem module and assess how long-horizon context affects navigation performance. As summarized in Tab.~\ref{tab:ablation_memory}, removing the memory module entirely (``w/o Memory'') leads to a substantial performance drop. This is particularly pronounced in large or multi-room layouts, where agents frequently get stuck in loops or revisit dead ends.
Using a naive replay buffer that stores only visual features (``Visual-only'') or only linguistic memory (``Linguistic-only'') partially recovers performance but remains inferior to our full approach.
In contrast, our proposed VLingMem achieves the best results with minimal latency overhead. Qualitatively, VLingMem enables the agent to remember the environment layout and avoid revisiting explored regions, yielding higher success rates with more efficient paths.


\subsubsection{\textbf{Co-train with Open-world Video Data}}
\begin{table}[htbp]
    \centering
    \small
    \caption{Ablation study on open-world video co-training.}
    \label{tab:ablation_cotrain}
    
    \scalebox{1}{
        \setlength{\tabcolsep}{10pt}
        \begin{tabular}{l cc ccc cc}
            \toprule
            \multirow{2}{*}{\textbf{Training Data}} & \multicolumn{2}{c}{\textbf{ObjNav}} & \multicolumn{3}{c}{\textbf{Track}} & \multicolumn{2}{c}{\textbf{ImageNav}} \\
            \cmidrule(lr){2-3} \cmidrule(lr){4-6} \cmidrule(lr){7-8}
             & SR$\uparrow$ & SPL$\uparrow$ & SR$\uparrow$ & TR$\uparrow$ & CR$\downarrow$ & SR$\uparrow$ & SPL$\uparrow$ \\
            \midrule
            
            w/o Co-training & 43.1 & 20.6 & 66.5 & 70.2 & 7.62 & 50.2 & 32.7 \\
            
            \rowcolor{gray!20}
            \textbf{w/ Co-training} & \textbf{50.1} & \textbf{24.6} & \textbf{67.6} & \textbf{73.5} & \textbf{5.51} & \textbf{60.8} & \textbf{37.4} \\
            \bottomrule
        \end{tabular}
    }
\end{table}
We further evaluated the impact of co-training with open-world video data. As presented in Tab.~\ref{tab:ablation_cotrain}, the results demonstrate a significant performance improvement compared to the model trained solely on embodied navigation data. This co-training strategy effectively enriches the model’s semantic priors, thereby enhancing cross-modal grounding and generalization capabilities while notably reducing the sim-to-real gap.


\subsubsection{\textbf{SFT Training Steps}}
\begin{figure}[!t]
    \centering
    \begin{minipage}[t]{0.37\linewidth}
        \centering
        \includegraphics[width=\linewidth]{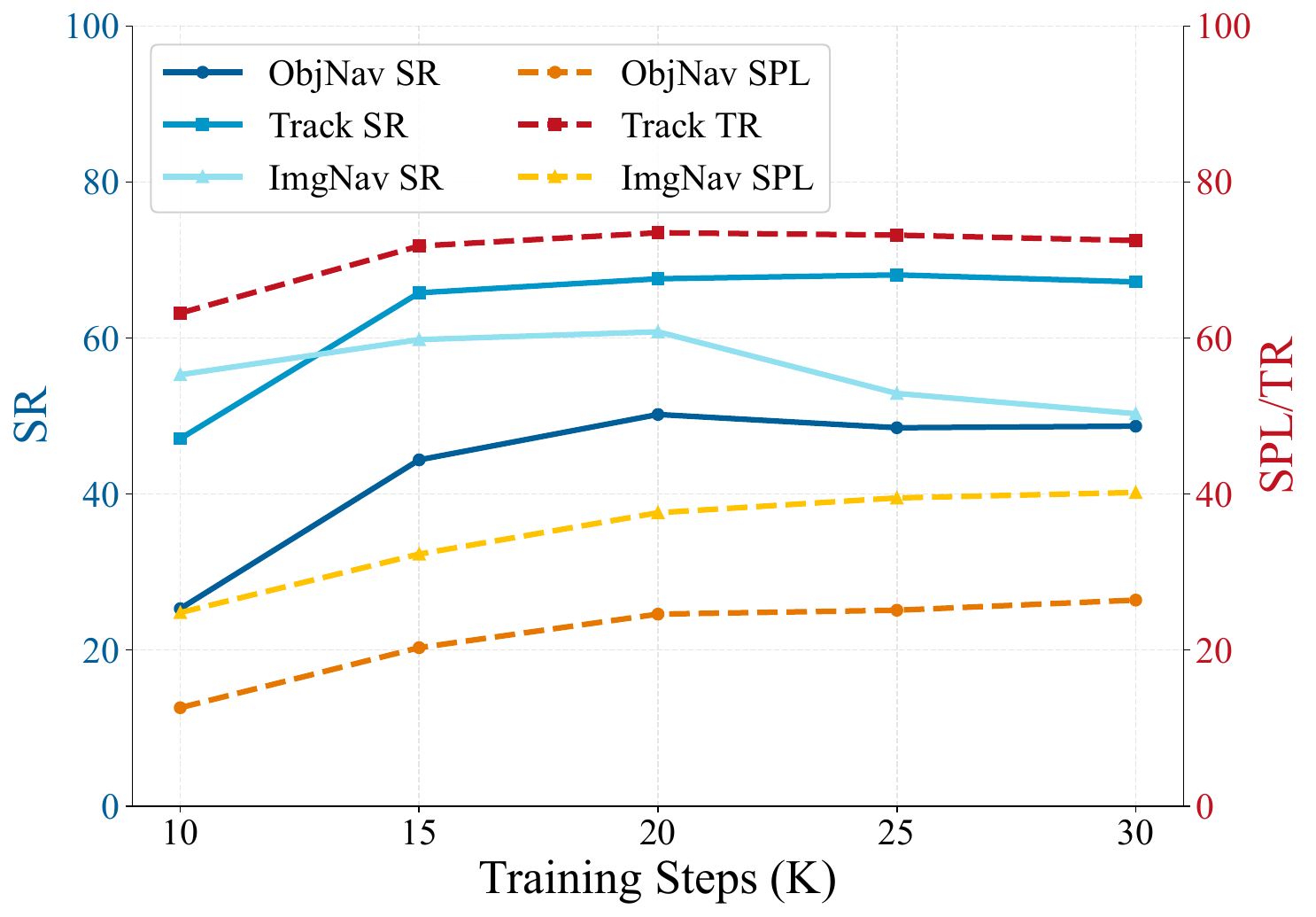}
        \caption{Ablation study on training steps.} 
        \label{fig:ablation1}
    \end{minipage}
    \begin{minipage}[t]{0.62\linewidth}
        \centering
        \includegraphics[width=\linewidth]{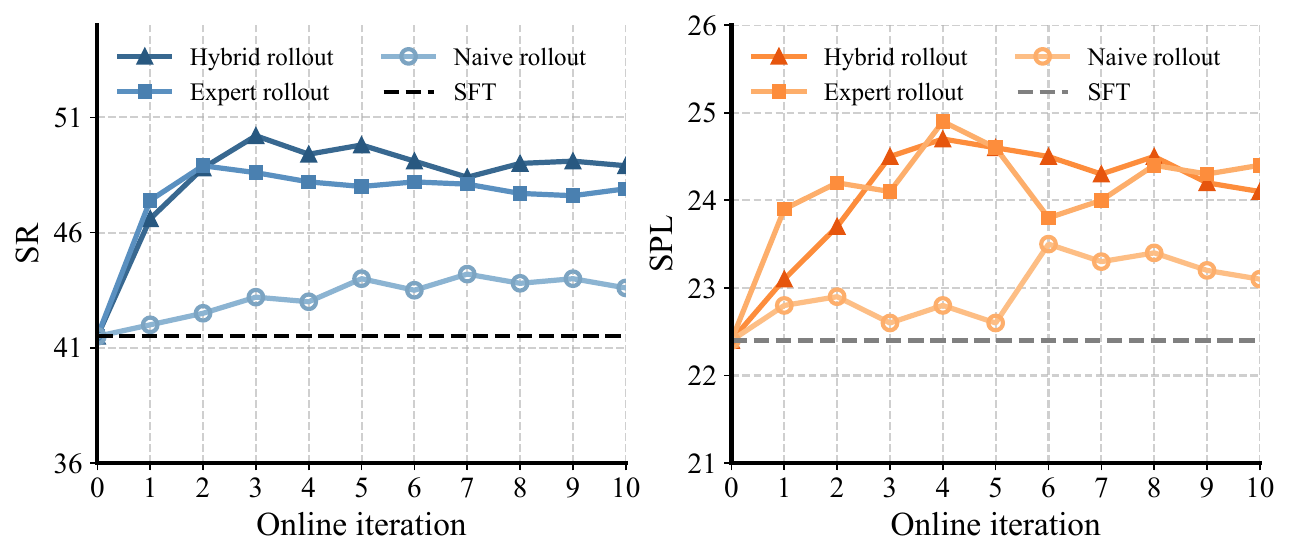}
        \caption{Ablation study on online post-training iteration steps.}
        \label{fig:ablation_post}
    \end{minipage}
\end{figure}

We investigated the relationship between model performance and the number of training steps. As shown in Fig.~\ref{fig:ablation1}, model performance scales positively with the number of training steps (where 1 epoch $\approx$ 10K training steps). The success rate rises steadily as the model is exposed to more data.
Notably, we found that excessive training leads to diminishing returns and eventual performance degradation, likely due to overfitting on the simulation data. This highlights the need for a balanced training strategy that maximizes performance without incurring unnecessary computational cost or risking overfitting.

\begin{figure}[!t]
    \centering
    \includegraphics[width=0.6\linewidth]{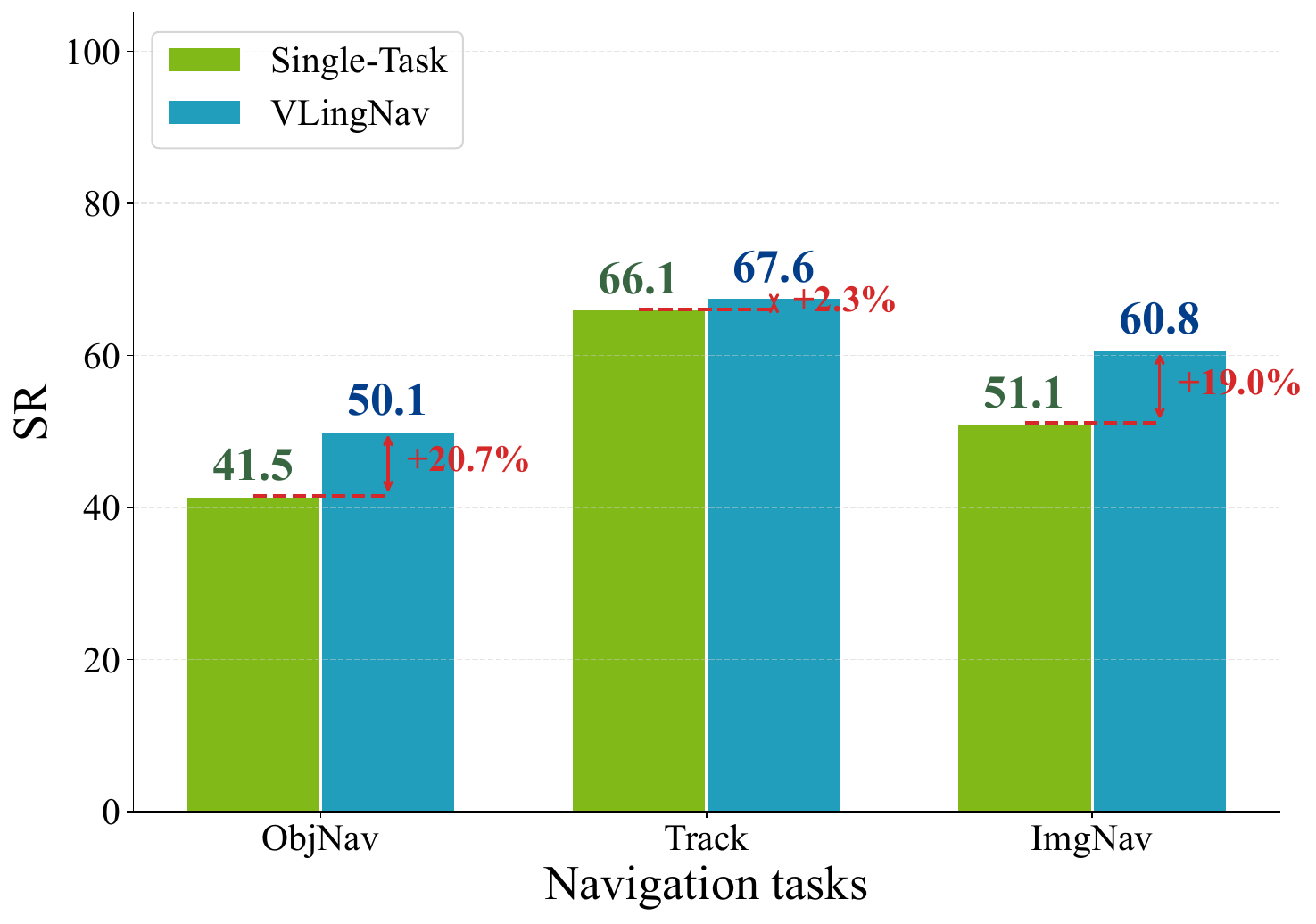}
    \caption{Ablation study on multi-task learning. We present the multi-task synergy of \ours and illustrate the performance comparison between models trained on a single task and those trained on multiple tasks.}
    \label{fig:ablation_multi}
\end{figure}

\subsubsection{\textbf{Online Post-training}}
We evaluated the effect of our online post-training phase, which follows the SFT stage. Across all benchmarks (Tab.~\ref{tab:objnav_hm3d},\ref{tab:objnav-ovon},\ref{tab:evt-bench},\ref{tab:imgnav}), the post-trained \ours model significantly outperforms the SFT checkpoint. This phase is critical for teaching the agent to find shortcuts, recover from errors, and handle the distribution shift that occurs beyond the static demonstration data.
As shown in Fig.~\ref{fig:ablation_post}, the ablation studies on the rollout strategy demonstrate that the proposed Hybrid Rollout exhibits the highest effectiveness, yielding the optimal performance. Here, we evaluate on the HM3D OVON \textit{val unseen} split. While the Expert Rollout (DAgger-like) also delivers strong performance, it still exhibits a performance gap compared to the Hybrid Rollout. However, the Naive Rollout fails to improve performance, likely due to the sparse reward signals and the long-horizon nature of the task making value estimation too difficult. This confirms that our expert-guided framework successfully optimizes the policy. It uses expert data to correct faulty behaviors while simultaneously using on-policy data to explore and discover better strategies, thereby outperforming pure imitation learning and finding a more robust policy.

\subsubsection{\textbf{Multi-task Synergy}}
Finally, we investigated how jointly training on ObjectNav, EVT, and ImageNav affects generalization. As shown in Fig.~\ref{fig:ablation_multi}, models trained on a single task consistently underperform the multi-task model, even on their respective specialized benchmarks.
More importantly, this multi-task training strategy fosters emergent cross-domain and cross-task capabilities, leading to a notable performance improvement on out-of-distribution tasks. These findings demonstrate that multi-task learning facilitates the transfer of skills across different domains. This synergy enhances the model’s ability to reason and plan across diverse modalities, tasks, and target categories.
\section{Discussion}
We further discuss the core contributions of our approach, their broader implications, and the key insights revealed by our experimental results.

\textbf{1) Effectiveness of Adaptive Thinking:} Inspired by dual-process theory, our adaptive Chain-of-Thought (AdaCoT) mechanism autonomously allocates ``cognitive resources'', balancing efficiency and deliberation.  When faced with simple, unambiguous navigation scenarios, such as traversing a straight corridor, the model opts for ``fast thinking'' (\texttt{<think\_off>}) and directly outputs actions, ensuring fluid and real-time navigation. Conversely, at critical decision positions, in complex environments, or when encountering ambiguity—like choosing a direction at an intersection or searching for an occluded object—the model triggers ``slow thinking'' (\texttt{<think\_on>}), generating a detailed reasoning output. This adaptability not only significantly enhances decision quality but also proves that deliberate thought at a small fraction of key steps (shown to be average 2.1\% in our experiments) is sufficient to substantially boost overall task success. This finding is crucial for deploying efficient, intelligent navigation on resource-constrained robot platforms.


\textbf{2) Synergy of Visual-Assisted Linguistic Memory:}
Navigation is a long-horizon decision-making process by its nature. Our proposed Visual-Assisted Linguistic Memory (VLingMem) module effectively addresses the memory deficiencies in traditional VLA models. Unlike methods that rely solely on implicit visual features, VLingMem distills key visual observations into concise linguistic summaries (\texttt{<summary>}$\cdots$\texttt{</summary>}) and integrates them into the model's context. This design offers two primary advantages. First, linguistic memory is more robust against information decay than compressed visual features, enabling the model to clearly recall critical semantic information such as ``I have already checked this room'' or ``there is a locked door on the left,'' thereby effectively preventing redundant exploration and inefficient paths. Second, this linguistic memory forms a powerful synergy with the AdaCoT mechanism. When the model chooses not to engage in detailed CoT, the persistent linguistic memory still provides the necessary historical context, ensuring a coherent decision-making process. This synergy serves as a pivotal factor in enhancing the robustness of \ours in long-horizon and complex environments, while significantly improving the efficiency and quality of \ours's exploration.

\textbf{3) Beyond Imitation Learning: The Value of Online Expert-guided RL:}
Our research confirms that VLA models trained exclusively via imitation learning (SFT) are constrained by both the quality and coverage of expert data. Such models are additionally prone to critical issues, including causal confusion and covariate shift. To address these limitations, we introduce a post-training phase using expert-guided reinforcement learning. \ours enables autonomous exploration and policy refinement through real-time online interaction with the environment, while directly deriving rewards from prior expert policy. Compared to rule-based RL, the introduction of expert knowledge allows the model to discover superior or more robust navigation strategies with higher efficiency. 
The significant performance gains observed in our experiments underscore that RL post-training is a critical step to unlock the full potential of VLA models, transforming them from mere ``imitators'' into genuine ``problem solvers.''

\textbf{4) Generality and Real-world Generalization:}
A notable achievement of \ours is its generality. By training on the large-scale, multi-task Nav-AdaCoT-2.9M dataset, \ours achieves state-of-the-art or competitive performance across all these tasks using a single, unified set of model weights, obviating the need for task-specific fine-tuning. This demonstrates that our approach successfully captures the underlying, universal cognitive structures of embodied navigation. Even more encouraging is \ours's ability to transfer to real-world robot platforms in a zero-shot manner and complete practical navigation tasks. This indicates that, through high-quality simulation training and a powerful cognitive architecture, the model learns generalizable representations of space, language, and action, rather than just patterns specific to the simulator, successfully bridging the sim-to-real gap.
In summary, \ours, with its unique cognitive architecture, provides a powerful paradigm for developing more intelligent, efficient, and interpretable embodied agents. It demonstrates the immense potential of combining principles from human cognition, such as adaptive thinking and episodic memory, with advanced machine learning paradigms like VLAs and RL.

\section{Conclusion and Limitation}
In this work, we introduce \ours, a Vision-Language-Action model grounded in linguistic-driven cognition to address critical challenges in embodied navigation. By synergistically integrating adaptive reasoning, multimodal memory, and online expert-guided RL post-training, \ours achieves state-of-the-art performance across a range of embodied navigation benchmarks and can directly transfer to real-world robot platforms in a zero-shot manner. 

While \ours has achieved progress in embodied navigation, it has several limitations that point to promising directions for future research. First, the current model primarily relies on monocular egocentric observations as input. Due to the limited field of view (FOV) inherent in monocular vision, such input constrains the model’s perceptual capabilities. Following recent work~\citep{zhang2025embodied}, we will explore integrating multi-view observations to improve navigation efficiency. Second, the current model adopts a single-system architecture, which restricts its prediction frequency. This limitation impedes rapid decision-making and obstacle handling in highly dynamic environments. To address this, we plan to upgrade \ours to a dual-system structure that supports high-frequency action outputs, thereby enhancing fundamental navigation performance, such as obstacle avoidance. Finally, the current approach uses only an MPC-based waypoint controller and lacks a more flexible locomotion model~\citep{miki2022learning}. Incorporating such a locomotion controller could increase movement speed and expand the robot’s reachable areas. We therefore plan to integrate locomotion capabilities into \ours in future work.

\section{Acknowledgements}
We sincerely thank Yunke Cai, Haiquan Chen, Shuai Chu, Taifeng Gao, Bo Jiang, Yunfei Li, Yunfei Liu, Tao Wang, Xibin Wu, and Tingshuai Yan for their strong support and fruitful discussions. 

\clearpage

\bibliographystyle{plainnat}
\bibliography{references}

\begin{thebibliography}{90}
\providecommand{\natexlab}[1]{#1}
\providecommand{\url}[1]{\texttt{#1}}
\expandafter\ifx\csname urlstyle\endcsname\relax
  \providecommand{\doi}[1]{doi: #1}\else
  \providecommand{\doi}{doi: \begingroup \urlstyle{rm}\Url}\fi

\bibitem[Azuma et~al.(2022)Azuma, Miyanishi, Kurita, and Kawanabe]{azuma2022scanqa}
Daichi Azuma, Taiki Miyanishi, Shuhei Kurita, and Motoaki Kawanabe.
\newblock Scanqa: 3d question answering for spatial scene understanding.
\newblock In \emph{proceedings of the IEEE/CVF conference on computer vision and pattern recognition}, pages 19129--19139, 2022.

\bibitem[Bai et~al.(2025)Bai, Chen, Liu, Wang, Ge, Song, Dang, Wang, Wang, Tang, et~al.]{bai2025qwen2}
Shuai Bai, Keqin Chen, Xuejing Liu, Jialin Wang, Wenbin Ge, Sibo Song, Kai Dang, Peng Wang, Shijie Wang, Jun Tang, et~al.
\newblock Qwen2. 5-vl technical report.
\newblock \emph{arXiv preprint arXiv:2502.13923}, 2025.

\bibitem[Black et~al.(2025)Black, Brown, Darpinian, Dhabalia, Driess, Esmail, Equi, Finn, Fusai, Galliker, et~al.]{black2025pi_}
Kevin Black, Noah Brown, James Darpinian, Karan Dhabalia, Danny Driess, Adnan Esmail, Michael~Robert Equi, Chelsea Finn, Niccolo Fusai, Manuel~Y Galliker, et~al.
\newblock $\pi_{0.5}$: a vision-language-action model with open-world generalization.
\newblock In \emph{9th Annual Conference on Robot Learning}, 2025.

\bibitem[Bolya et~al.(2022)Bolya, Fu, Dai, Zhang, Feichtenhofer, and Hoffman]{Bolya2022TokenMY}
Daniel Bolya, Cheng-Yang Fu, Xiaoliang Dai, Peizhao Zhang, Christoph Feichtenhofer, and Judy Hoffman.
\newblock Token merging: Your vit but faster.
\newblock \emph{arXiv preprint arXiv:2210.09461}, 2022.

\bibitem[Campos et~al.(2021)Campos, Elvira, Rodr{\'\i}guez, Montiel, and Tard{\'o}s]{campos2021orb}
Carlos Campos, Richard Elvira, Juan J~G{\'o}mez Rodr{\'\i}guez, Jos{\'e}~MM Montiel, and Juan~D Tard{\'o}s.
\newblock Orb-slam3: An accurate open-source library for visual, visual--inertial, and multimap slam.
\newblock \emph{IEEE transactions on robotics}, 37\penalty0 (6):\penalty0 1874--1890, 2021.

\bibitem[Cao et~al.(2025)Cao, Zhang, Yu, Liu, Qin, Zou, Du, and Xu]{cao2024cognav}
Yihan Cao, Jiazhao Zhang, Zhinan Yu, Shuzhen Liu, Zheng Qin, Qin Zou, Bo~Du, and Kai Xu.
\newblock Cognav: Cognitive process modeling for object goal navigation with llms.
\newblock In \emph{Proceedings of the IEEE/CVF International Conference on Computer Vision}, pages 9550--9560, 2025.

\bibitem[Chang et~al.(2017)Chang, Dai, Funkhouser, Halber, Niebner, Savva, Song, Zeng, and Zhang]{chang2017matterport3d}
Angel Chang, Angela Dai, Thomas Funkhouser, Maciej Halber, Matthias Niebner, Manolis Savva, Shuran Song, Andy Zeng, and Yinda Zhang.
\newblock Matterport3d: Learning from rgb-d data in indoor environments.
\newblock In \emph{2017 International Conference on 3D Vision (3DV)}, pages 667--676. IEEE, 2017.

\bibitem[Chang et~al.(2023)Chang, Gervet, Khanna, Yenamandra, Shah, Min, Shah, Paxton, Gupta, Batra, et~al.]{chang2023goat}
Matthew Chang, Theophile Gervet, Mukul Khanna, Sriram Yenamandra, Dhruv Shah, So~Yeon Min, Kavit Shah, Chris Paxton, Saurabh Gupta, Dhruv Batra, et~al.
\newblock Goat: Go to any thing.
\newblock \emph{arXiv preprint arXiv:2311.06430}, 2023.

\bibitem[Chen et~al.(2025)Chen, He, Hu, Liu, Wang, Xu, Zhang, Zhang, An, Cai, et~al.]{chen2025astra}
Sheng Chen, Peiyu He, Jiaxin Hu, Ziyang Liu, Yansheng Wang, Tao Xu, Chi Zhang, Chongchong Zhang, Chao An, Shiyu Cai, et~al.
\newblock Astra: Toward general-purpose mobile robots via hierarchical multimodal learning.
\newblock \emph{arXiv preprint arXiv:2506.06205}, 2025.

\bibitem[Cheng et~al.(2025)Cheng, Ji, Yang, Zou, Kautz, Biyik, Yin, Liu, and Wang]{cheng2024navila}
An-Chieh Cheng, Yandong Ji, Zhaojing Yang, Xueyan Zou, Jan Kautz, Erdem Biyik, Hongxu Yin, Sifei Liu, and Xiaolong Wang.
\newblock Navila: Legged robot vision-language-action model for navigation.
\newblock In \emph{RSS}, 2025.

\bibitem[DeSouza and Kak(2002)]{desouza2002vision}
Guilherme~N DeSouza and Avinash~C Kak.
\newblock Vision for mobile robot navigation: A survey.
\newblock \emph{IEEE transactions on pattern analysis and machine intelligence}, 24\penalty0 (2):\penalty0 237--267, 2002.

\bibitem[Ebbinghaus(2013)]{ebbinghaus2013image}
Hermann Ebbinghaus.
\newblock [image] memory: A contribution to experimental psychology.
\newblock \emph{Annals of neurosciences}, 20\penalty0 (4):\penalty0 155, 2013.

\bibitem[Feng et~al.(2025)Feng, Gong, Li, Guo, Wang, Peng, Wu, Zhang, Wang, and Yue]{feng2025video}
Kaituo Feng, Kaixiong Gong, Bohao Li, Zonghao Guo, Yibing Wang, Tianshuo Peng, Junfei Wu, Xiaoying Zhang, Benyou Wang, and Xiangyu Yue.
\newblock Video-r1: Reinforcing video reasoning in mllms.
\newblock \emph{arXiv preprint arXiv:2503.21776}, 2025.

\bibitem[Gao et~al.(2025)Gao, Jin, Peng, Zhang, Deng, Li, Wang, and Liu]{gao2025octonav}
Chen Gao, Liankai Jin, Xingyu Peng, Jiazhao Zhang, Yue Deng, Annan Li, He~Wang, and Si~Liu.
\newblock Octonav: Towards generalist embodied navigation.
\newblock \emph{arXiv preprint arXiv:2506.09839}, 2025.

\bibitem[Grandia et~al.(2023)Grandia, Jenelten, Yang, Farshidian, and Hutter]{grandia2023perceptive}
Ruben Grandia, Fabian Jenelten, Shao Yang, Farbod Farshidian, and Marco Hutter.
\newblock Perceptive locomotion through nonlinear model predictive control.
\newblock \emph{IEEE Transactions on Robotics}, 39\penalty0 (5):\penalty0 3402--3421, 2023.

\bibitem[Guo et~al.(2025)Guo, Yang, Zhang, Song, Zhang, Xu, Zhu, Ma, Wang, Bi, et~al.]{guo2025deepseek}
Daya Guo, Dejian Yang, Haowei Zhang, Junxiao Song, Ruoyu Zhang, Runxin Xu, Qihao Zhu, Shirong Ma, Peiyi Wang, Xiao Bi, et~al.
\newblock Deepseek-r1: Incentivizing reasoning capability in llms via reinforcement learning.
\newblock \emph{arXiv preprint arXiv:2501.12948}, 2025.

\bibitem[Gupta et~al.(2016)Gupta, Kumar, Behera, and Subramanian]{gupta2016novel}
Meenakshi Gupta, Swagat Kumar, Laxmidhar Behera, and Venkatesh~K Subramanian.
\newblock A novel vision-based tracking algorithm for a human-following mobile robot.
\newblock \emph{IEEE Transactions on Systems, Man, and Cybernetics: Systems}, 47\penalty0 (7):\penalty0 1415--1427, 2016.

\bibitem[Hu(2025)]{hu2025reinforce++}
Jian Hu.
\newblock Reinforce++: A simple and efficient approach for aligning large language models.
\newblock \emph{arXiv preprint arXiv:2501.03262}, 2025.

\bibitem[Hu et~al.()Hu, Guo, Wang, Chen, Wang, Zhang, Sreenath, Lu, and Chen]{hu2024video}
Yucheng Hu, Yanjiang Guo, Pengchao Wang, Xiaoyu Chen, Yen-Jen Wang, Jianke Zhang, Koushil Sreenath, Chaochao Lu, and Jianyu Chen.
\newblock Video prediction policy: A generalist robot policy with predictive visual representations.
\newblock In \emph{Forty-second International Conference on Machine Learning}.

\bibitem[Huang et~al.(2025)Huang, Wu, Chen, Wang, and Yang]{huang2025thinkact}
Chi-Pin Huang, Yueh-Hua Wu, Min-Hung Chen, Yu-Chiang~Frank Wang, and Fu-En Yang.
\newblock Thinkact: Vision-language-action reasoning via reinforced visual latent planning.
\newblock \emph{arXiv preprint arXiv:2507.16815}, 2025.

\bibitem[Karaman et~al.(2011)Karaman, Walter, Perez, Frazzoli, and Teller]{karaman2011anytime}
Sertac Karaman, Matthew~R Walter, Alejandro Perez, Emilio Frazzoli, and Seth Teller.
\newblock Anytime motion planning using the rrt.
\newblock In \emph{2011 IEEE International Conference on Robotics and Automation}, pages 1478--1483. ieee, 2011.

\bibitem[Kavraki et~al.(2002)Kavraki, Svestka, Latombe, and Overmars]{kavraki2002probabilistic}
Lydia~E Kavraki, Petr Svestka, J-C Latombe, and Mark~H Overmars.
\newblock Probabilistic roadmaps for path planning in high-dimensional configuration spaces.
\newblock \emph{IEEE Transactions on Robotics and Automation}, 12\penalty0 (4):\penalty0 566--580, 2002.

\bibitem[Kim et~al.(2025)Kim, Finn, and Liang]{kim2025fine}
Moo~Jin Kim, Chelsea Finn, and Percy Liang.
\newblock Fine-tuning vision-language-action models: Optimizing speed and success.
\newblock \emph{arXiv preprint arXiv:2502.19645}, 2025.

\bibitem[Krantz et~al.(2022)Krantz, Lee, Malik, Batra, and Chaplot]{krantz2022instance}
Jacob Krantz, Stefan Lee, Jitendra Malik, Dhruv Batra, and Devendra~Singh Chaplot.
\newblock Instance-specific image goal navigation: Training embodied agents to find object instances.
\newblock \emph{arXiv preprint arXiv:2211.15876}, 2022.

\bibitem[Krantz et~al.(2023)Krantz, Gervet, Yadav, Wang, Paxton, Mottaghi, Batra, Malik, Lee, and Chaplot]{krantz2023navigating}
Jacob Krantz, Theophile Gervet, Karmesh Yadav, Austin Wang, Chris Paxton, Roozbeh Mottaghi, Dhruv Batra, Jitendra Malik, Stefan Lee, and Devendra~Singh Chaplot.
\newblock Navigating to objects specified by images.
\newblock In \emph{Proceedings of the IEEE/CVF International Conference on Computer Vision}, pages 10916--10925, 2023.

\bibitem[Li et~al.(2025)Li, Zuo, Yu, Zhang, Yang, Zhang, Zhu, Zhang, Chen, Cui, et~al.]{li2025simplevla}
Haozhan Li, Yuxin Zuo, Jiale Yu, Yuhao Zhang, Zhaohui Yang, Kaiyan Zhang, Xuekai Zhu, Yuchen Zhang, Tianxing Chen, Ganqu Cui, et~al.
\newblock Simplevla-rl: Scaling vla training via reinforcement learning.
\newblock \emph{arXiv preprint arXiv:2509.09674}, 2025.

\bibitem[Li et~al.(2023)Li, Liu, Zhang, Yu, Xu, Wu, Cheang, Jing, Zhang, Liu, et~al.]{li2023vision}
Xinghang Li, Minghuan Liu, Hanbo Zhang, Cunjun Yu, Jie Xu, Hongtao Wu, Chilam Cheang, Ya~Jing, Weinan Zhang, Huaping Liu, et~al.
\newblock Vision-language foundation models as effective robot imitators.
\newblock \emph{arXiv preprint arXiv:2311.01378}, 2023.

\bibitem[Lindenberger et~al.(2023)Lindenberger, Sarlin, and Pollefeys]{lindenberger2023lightglue}
Philipp Lindenberger, Paul-Edouard Sarlin, and Marc Pollefeys.
\newblock Lightglue: Local feature matching at light speed.
\newblock In \emph{Proceedings of the IEEE/CVF international conference on computer vision}, pages 17627--17638, 2023.

\bibitem[Liu et~al.(2023)Liu, Li, Wu, and Lee]{liu2023llava}
Haotian Liu, Chunyuan Li, Qingyang Wu, and Yong~Jae Lee.
\newblock Visual instruction tuning.
\newblock In \emph{NeurIPS}, 2023.

\bibitem[Liu et~al.(2025{\natexlab{a}})Liu, Qi, Zhang, Li, Wang, Wu, Ye, Zhang, Chen, Zhong, et~al.]{liu2025trackvla++}
Jiahang Liu, Yunpeng Qi, Jiazhao Zhang, Minghan Li, Shaoan Wang, Kui Wu, Hanjing Ye, Hong Zhang, Zhibo Chen, Fangwei Zhong, et~al.
\newblock Trackvla++: Unleashing reasoning and memory capabilities in vla models for embodied visual tracking.
\newblock \emph{arXiv preprint arXiv:2510.07134}, 2025{\natexlab{a}}.

\bibitem[Liu et~al.(2025{\natexlab{b}})Liu, Huang, Zhang, and Tang]{liu2025nav}
Qingxiang Liu, Ting Huang, Zeyu Zhang, and Hao Tang.
\newblock Nav-r1: Reasoning and navigation in embodied scenes.
\newblock \emph{arXiv preprint arXiv:2509.10884}, 2025{\natexlab{b}}.

\bibitem[Liu et~al.(2024)Liu, Zeng, Ren, Li, Zhang, Yang, Jiang, Li, Yang, Su, et~al.]{liu2023grounding}
Shilong Liu, Zhaoyang Zeng, Tianhe Ren, Feng Li, Hao Zhang, Jie Yang, Qing Jiang, Chunyuan Li, Jianwei Yang, Hang Su, et~al.
\newblock Grounding dino: Marrying dino with grounded pre-training for open-set object detection.
\newblock In \emph{European conference on computer vision}, pages 38--55. Springer, 2024.

\bibitem[Long et~al.(2024)Long, Cai, Wang, Zhan, and Dong]{long2024instructnav}
Yuxing Long, Wenzhe Cai, Hongcheng Wang, Guanqi Zhan, and Hao Dong.
\newblock Instructnav: Zero-shot system for generic instruction navigation in unexplored environment.
\newblock \emph{arXiv preprint arXiv:2406.04882}, 2024.

\bibitem[Luo et~al.(2018)Luo, Sun, Zhong, Liu, Zhang, and Wang]{luo2018end}
Wenhan Luo, Peng Sun, Fangwei Zhong, Wei Liu, Tong Zhang, and Yizhou Wang.
\newblock End-to-end active object tracking via reinforcement learning.
\newblock In \emph{International conference on machine learning}, pages 3286--3295. PMLR, 2018.

\bibitem[Miki et~al.(2022)Miki, Lee, Hwangbo, Wellhausen, Koltun, and Hutter]{miki2022learning}
Takahiro Miki, Joonho Lee, Jemin Hwangbo, Lorenz Wellhausen, Vladlen Koltun, and Marco Hutter.
\newblock Learning robust perceptive locomotion for quadrupedal robots in the wild.
\newblock \emph{Science robotics}, 7\penalty0 (62):\penalty0 eabk2822, 2022.

\bibitem[OpenAI(2024)]{openai2024introducing}
OpenAI.
\newblock Introducing 4o image generation.
\newblock \url{https://openai.com/index/introducing-4o-image}, 2024.
\newblock Accessed: 2025-04-29.

\bibitem[Qi et~al.(2025)Qi, Zhang, Yu, Wang, and Zhao]{qi2025vln}
Zhangyang Qi, Zhixiong Zhang, Yizhou Yu, Jiaqi Wang, and Hengshuang Zhao.
\newblock Vln-r1: Vision-language navigation via reinforcement fine-tuning.
\newblock \emph{arXiv preprint arXiv:2506.17221}, 2025.

\bibitem[Qin et~al.(2018)Qin, Li, and Shen]{qin2018vins}
Tong Qin, Peiliang Li, and Shaojie Shen.
\newblock Vins-mono: A robust and versatile monocular visual-inertial state estimator.
\newblock \emph{IEEE Transactions on Robotics}, 34\penalty0 (4):\penalty0 1004--1020, 2018.

\bibitem[Radford et~al.(2021)Radford, Kim, Hallacy, Ramesh, Goh, Agarwal, Sastry, Askell, Mishkin, Clark, et~al.]{radford2021clip}
Alec Radford, Jong~Wook Kim, Chris Hallacy, Aditya Ramesh, Gabriel Goh, Sandhini Agarwal, Girish Sastry, Amanda Askell, Pamela Mishkin, Jack Clark, et~al.
\newblock Learning transferable visual models from natural language supervision.
\newblock In \emph{International Conference on Machine Learning}, pages 8748--8763. PMLR, 2021.

\bibitem[Rajeswaran et~al.(2017)Rajeswaran, Kumar, Gupta, Vezzani, Schulman, Todorov, and Levine]{rajeswaran2017learning}
Aravind Rajeswaran, Vikash Kumar, Abhishek Gupta, Giulia Vezzani, John Schulman, Emanuel Todorov, and Sergey Levine.
\newblock Learning complex dexterous manipulation with deep reinforcement learning and demonstrations.
\newblock \emph{arXiv preprint arXiv:1709.10087}, 2017.

\bibitem[Ramakrishnan et~al.(2021)Ramakrishnan, Gokaslan, Wijmans, Maksymets, Clegg, Turner, Undersander, Galuba, Westbury, Chang, et~al.]{ramakrishnan2021habitat}
Santhosh~K Ramakrishnan, Aaron Gokaslan, Erik Wijmans, Oleksandr Maksymets, Alex Clegg, John Turner, Eric Undersander, Wojciech Galuba, Andrew Westbury, Angel~X Chang, et~al.
\newblock Habitat-matterport 3d dataset (hm3d): 1000 large-scale 3d environments for embodied ai.
\newblock \emph{arXiv preprint arXiv:2109.08238}, 2021.

\bibitem[Ramrakhya et~al.(2022)Ramrakhya, Undersander, Batra, and Das]{ramrakhya2022habitat}
Ram Ramrakhya, Eric Undersander, Dhruv Batra, and Abhishek Das.
\newblock Habitat-web: Learning embodied object-search strategies from human demonstrations at scale.
\newblock In \emph{Proceedings of the IEEE/CVF Conference on Computer Vision and Pattern Recognition}, pages 5173--5183, 2022.

\bibitem[Ramrakhya et~al.(2023)Ramrakhya, Batra, Wijmans, and Das]{ramrakhya2023pirlnav}
Ram Ramrakhya, Dhruv Batra, Erik Wijmans, and Abhishek Das.
\newblock Pirlnav: Pretraining with imitation and rl finetuning for objectnav.
\newblock In \emph{Proceedings of the IEEE/CVF Conference on Computer Vision and Pattern Recognition}, pages 17896--17906, 2023.

\bibitem[Ravi et~al.(2024)Ravi, Gabeur, Hu, Hu, Ryali, Ma, Khedr, R{\"a}dle, Rolland, Gustafson, et~al.]{ravi2024sam}
Nikhila Ravi, Valentin Gabeur, Yuan-Ting Hu, Ronghang Hu, Chaitanya Ryali, Tengyu Ma, Haitham Khedr, Roman R{\"a}dle, Chloe Rolland, Laura Gustafson, et~al.
\newblock Sam 2: Segment anything in images and videos.
\newblock \emph{arXiv preprint arXiv:2408.00714}, 2024.

\bibitem[Schulman et~al.(2017)Schulman, Wolski, Dhariwal, Radford, and Klimov]{schulman2017proximal}
John Schulman, Filip Wolski, Prafulla Dhariwal, Alec Radford, and Oleg Klimov.
\newblock Proximal policy optimization algorithms.
\newblock \emph{arXiv preprint arXiv:1707.06347}, 2017.

\bibitem[Shah et~al.(2023)Shah, Equi, Osi{\'n}ski, Xia, Ichter, and Levine]{shah2023navigation}
Dhruv Shah, Michael~Robert Equi, B{\l}a{\.z}ej Osi{\'n}ski, Fei Xia, Brian Ichter, and Sergey Levine.
\newblock Navigation with large language models: Semantic guesswork as a heuristic for planning.
\newblock In \emph{Conference on Robot Learning}, pages 2683--2699. PMLR, 2023.

\bibitem[Shao et~al.(2024)Shao, Wang, Zhu, Xu, Song, Bi, Zhang, Zhang, Li, Wu, et~al.]{shao2024deepseekmath}
Zhihong Shao, Peiyi Wang, Qihao Zhu, Runxin Xu, Junxiao Song, Xiao Bi, Haowei Zhang, Mingchuan Zhang, YK~Li, Yang Wu, et~al.
\newblock Deepseekmath: Pushing the limits of mathematical reasoning in open language models.
\newblock \emph{arXiv preprint arXiv:2402.03300}, 2024.

\bibitem[Shi et~al.(2025)Shi, Xie, Liu, Sun, Liu, Wang, Zhou, Fan, Zhang, and Huang]{shi2025memoryvla}
Hao Shi, Bin Xie, Yingfei Liu, Lin Sun, Fengrong Liu, Tiancai Wang, Erjin Zhou, Haoqiang Fan, Xiangyu Zhang, and Gao Huang.
\newblock Memoryvla: Perceptual-cognitive memory in vision-language-action models for robotic manipulation.
\newblock \emph{arXiv preprint arXiv:2508.19236}, 2025.

\bibitem[Su et~al.(2024)Su, Ahmed, Lu, Pan, Bo, and Liu]{su2024roformer}
Jianlin Su, Murtadha Ahmed, Yu~Lu, Shengfeng Pan, Wen Bo, and Yunfeng Liu.
\newblock Roformer: Enhanced transformer with rotary position embedding.
\newblock \emph{Neurocomputing}, 568:\penalty0 127063, 2024.

\bibitem[Sun et~al.(2024)Sun, Liu, Zhi, Qiu, and Liang]{sun2024prioritized}
Xinyu Sun, Lizhao Liu, Hongyan Zhi, Ronghe Qiu, and Junwei Liang.
\newblock Prioritized semantic learning for zero-shot instance navigation.
\newblock In \emph{European Conference on Computer Vision}, pages 161--178. Springer, 2024.

\bibitem[Wang et~al.(2025{\natexlab{a}})Wang, Chen, Karaev, Vedaldi, Rupprecht, and Novotny]{wang2025vggt}
Jianyuan Wang, Minghao Chen, Nikita Karaev, Andrea Vedaldi, Christian Rupprecht, and David Novotny.
\newblock Vggt: Visual geometry grounded transformer.
\newblock In \emph{Proceedings of the Computer Vision and Pattern Recognition Conference}, pages 5294--5306, 2025{\natexlab{a}}.

\bibitem[Wang et~al.(2025{\natexlab{b}})Wang, Zhang, Li, Liu, Li, Wu, Zhong, Yu, Zhang, and Wang]{wang2025trackvla}
Shaoan Wang, Jiazhao Zhang, Minghan Li, Jiahang Liu, Anqi Li, Kui Wu, Fangwei Zhong, Junzhi Yu, Zhizheng Zhang, and He~Wang.
\newblock Trackvla: Embodied visual tracking in the wild.
\newblock \emph{arXiv preprint arXiv:2505.23189}, 2025{\natexlab{b}}.

\bibitem[Wang et~al.(2025{\natexlab{c}})Wang, Wang, Li, Cai, Wang, Chen, Wang, Su, Li, and Fan]{wang2025think}
Shuo Wang, Yongcai Wang, Wanting Li, Xudong Cai, Yucheng Wang, Maiyue Chen, Kaihui Wang, Zhizhong Su, Deying Li, and Zhaoxin Fan.
\newblock Aux-think: Exploring reasoning strategies for data-efficient vision-language navigation.
\newblock \emph{arXiv preprint arXiv:2505.11886}, 2025{\natexlab{c}}.

\bibitem[Wang and Zhou(2024)]{wang2024chain}
Xuezhi Wang and Denny Zhou.
\newblock Chain-of-thought reasoning without prompting.
\newblock \emph{Advances in Neural Information Processing Systems}, 37:\penalty0 66383--66409, 2024.

\bibitem[Wei et~al.(2022)Wei, Wang, Schuurmans, Bosma, Xia, Chi, Le, Zhou, et~al.]{wei2022chain}
Jason Wei, Xuezhi Wang, Dale Schuurmans, Maarten Bosma, Fei Xia, Ed~Chi, Quoc~V Le, Denny Zhou, et~al.
\newblock Chain-of-thought prompting elicits reasoning in large language models.
\newblock \emph{Advances in Neural Information Processing Systems}, 35:\penalty0 24824--24837, 2022.

\bibitem[Wei et~al.(2025)Wei, Wan, Yu, Wang, Yang, Mao, Zhu, Cai, Wang, Chen, et~al.]{wei2025streamvln}
Meng Wei, Chenyang Wan, Xiqian Yu, Tai Wang, Yuqiang Yang, Xiaohan Mao, Chenming Zhu, Wenzhe Cai, Hanqing Wang, Yilun Chen, et~al.
\newblock Streamvln: Streaming vision-and-language navigation via slowfast context modeling.
\newblock \emph{arXiv preprint arXiv:2507.05240}, 2025.

\bibitem[Wu et~al.(2024)Wu, Zhang, Gu, Zheng, and Bai]{wu2024embodied}
Yuchen Wu, Pengcheng Zhang, Meiying Gu, Jin Zheng, and Xiao Bai.
\newblock Embodied navigation with multi-modal information: A survey from tasks to methodology.
\newblock \emph{Information Fusion}, 113:\penalty0 102532, 2024.

\bibitem[Xu et~al.(2022)Xu, Cai, He, Lin, and Zhang]{xu2022fast}
Wei Xu, Yixi Cai, Dongjiao He, Jiarong Lin, and Fu~Zhang.
\newblock Fast-lio2: Fast direct lidar-inertial odometry.
\newblock \emph{IEEE Transactions on Robotics}, 38\penalty0 (4):\penalty0 2053--2073, 2022.

\bibitem[Yadav et~al.(2023{\natexlab{a}})Yadav, Majumdar, Ramrakhya, Yokoyama, Baevski, Kira, Maksymets, and Batra]{yadav2023ovrl}
Karmesh Yadav, Arjun Majumdar, Ram Ramrakhya, Naoki Yokoyama, Alexei Baevski, Zsolt Kira, Oleksandr Maksymets, and Dhruv Batra.
\newblock Ovrl-v2: A simple state-of-art baseline for imagenav and objectnav.
\newblock \emph{arXiv preprint arXiv:2303.07798}, 2023{\natexlab{a}}.

\bibitem[Yadav et~al.(2023{\natexlab{b}})Yadav, Ramrakhya, Majumdar, Berges, Kuhar, Batra, Baevski, and Maksymets]{yadav2023offline}
Karmesh Yadav, Ram Ramrakhya, Arjun Majumdar, Vincent-Pierre Berges, Sachit Kuhar, Dhruv Batra, Alexei Baevski, and Oleksandr Maksymets.
\newblock Offline visual representation learning for embodied navigation.
\newblock In \emph{Workshop on Reincarnating Reinforcement Learning at ICLR 2023}, 2023{\natexlab{b}}.

\bibitem[Yang et~al.(2023)Yang, Zhang, Li, Zou, Li, and Gao]{yang2023set}
Jianwei Yang, Hao Zhang, Feng Li, Xueyan Zou, Chunyuan Li, and Jianfeng Gao.
\newblock Set-of-mark prompting unleashes extraordinary visual grounding in gpt-4v.
\newblock \emph{arXiv preprint arXiv:2310.11441}, 2023.

\bibitem[Yin et~al.(2024)Yin, Xu, Wu, Zhou, and Lu]{yin2024sg}
Hang Yin, Xiuwei Xu, Zhenyu Wu, Jie Zhou, and Jiwen Lu.
\newblock Sg-nav: Online 3d scene graph prompting for llm-based zero-shot object navigation.
\newblock \emph{Advances in neural information processing systems}, 37:\penalty0 5285--5307, 2024.

\bibitem[Yin et~al.(2025)Yin, Xu, Zhao, Wang, Zhou, and Lu]{yin2025unigoal}
Hang Yin, Xiuwei Xu, Linqing Zhao, Ziwei Wang, Jie Zhou, and Jiwen Lu.
\newblock Unigoal: Towards universal zero-shot goal-oriented navigation.
\newblock In \emph{Proceedings of the Computer Vision and Pattern Recognition Conference}, pages 19057--19066, 2025.

\bibitem[Yokoyama and Ha(2025)]{yokoyama2025film}
Naoki Yokoyama and Sehoon Ha.
\newblock Film-nav: Efficient and generalizable navigation via vlm fine-tuning.
\newblock \emph{arXiv preprint arXiv:2509.16445}, 2025.

\bibitem[Yokoyama et~al.(2024{\natexlab{a}})Yokoyama, Ha, Batra, Wang, and Bucher]{yokoyama2024vlfm}
Naoki Yokoyama, Sehoon Ha, Dhruv Batra, Jiuguang Wang, and Bernadette Bucher.
\newblock Vlfm: Vision-language frontier maps for zero-shot semantic navigation.
\newblock In \emph{2024 IEEE International Conference on Robotics and Automation (ICRA)}, pages 42--48. IEEE, 2024{\natexlab{a}}.

\bibitem[Yokoyama et~al.(2024{\natexlab{b}})Yokoyama, Ramrakhya, Das, Batra, and Ha]{yokoyama2024hm3d}
Naoki Yokoyama, Ram Ramrakhya, Abhishek Das, Dhruv Batra, and Sehoon Ha.
\newblock Hm3d-ovon: A dataset and benchmark for open-vocabulary object goal navigation.
\newblock In \emph{2024 IEEE/RSJ International Conference on Intelligent Robots and Systems (IROS)}, pages 5543--5550. IEEE, 2024{\natexlab{b}}.

\bibitem[Yu et~al.(2023)Yu, Kasaei, and Cao]{yu2023l3mvn}
Bangguo Yu, Hamidreza Kasaei, and Ming Cao.
\newblock L3mvn: Leveraging large language models for visual target navigation.
\newblock In \emph{2023 IEEE/RSJ International Conference on Intelligent Robots and Systems (IROS)}, pages 3554--3560. IEEE, 2023.

\bibitem[Zawalski et~al.(2024)Zawalski, Chen, Pertsch, Mees, Finn, and Levine]{zawalski2024robotic}
Micha{\l} Zawalski, William Chen, Karl Pertsch, Oier Mees, Chelsea Finn, and Sergey Levine.
\newblock Robotic control via embodied chain-of-thought reasoning.
\newblock \emph{arXiv preprint arXiv:2407.08693}, 2024.

\bibitem[Zeng et~al.()Zeng, Zhang, Ehsani, Hendrix, Salvador, Herrasti, Girshick, Kembhavi, and Weihs]{zeng2024poliformer}
Kuo-Hao Zeng, Zichen Zhang, Kiana Ehsani, Rose Hendrix, Jordi Salvador, Alvaro Herrasti, Ross Girshick, Aniruddha Kembhavi, and Luca Weihs.
\newblock Poliformer: Scaling on-policy rl with transformers results in masterful navigators.
\newblock In \emph{8th Annual Conference on Robot Learning}.

\bibitem[Zeng et~al.(2025)Zeng, Qi, Chang, Xiong, Xie, Wu, Liang, Xu, and Wei]{zeng2025janusvln}
Shuang Zeng, Dekang Qi, Xinyuan Chang, Feng Xiong, Shichao Xie, Xiaolong Wu, Shiyi Liang, Mu~Xu, and Xing Wei.
\newblock Janusvln: Decoupling semantics and spatiality with dual implicit memory for vision-language navigation.
\newblock \emph{arXiv preprint arXiv:2509.22548}, 2025.

\bibitem[Zhai et~al.(2023)Zhai, Mustafa, Kolesnikov, and Beyer]{zhai2023siglip}
Xiaohua Zhai, Basil Mustafa, Alexander Kolesnikov, and Lucas Beyer.
\newblock Sigmoid loss for language image pre-training.
\newblock In \emph{Proceedings of the IEEE/CVF international conference on computer vision}, pages 11975--11986, 2023.

\bibitem[Zhang et~al.()Zhang, Guo, Hu, Chen, Zhu, and Chen]{zhang2024up}
Jianke Zhang, Yanjiang Guo, Yucheng Hu, Xiaoyu Chen, Xiang Zhu, and Jianyu Chen.
\newblock Up-vla: A unified understanding and prediction model for embodied agent.
\newblock In \emph{Forty-second International Conference on Machine Learning}.

\bibitem[Zhang et~al.(2024{\natexlab{a}})Zhang, Wang, Xu, Zhou, Hong, Fang, Wu, Zhang, and Wang]{zhang2024navid}
Jiazhao Zhang, Kunyu Wang, Rongtao Xu, Gengze Zhou, Yicong Hong, Xiaomeng Fang, Qi~Wu, Zhizheng Zhang, and He~Wang.
\newblock Navid: Video-based vlm plans the next step for vision-and-language navigation.
\newblock \emph{Robotics: Science and Systems}, 2024{\natexlab{a}}.

\bibitem[Zhang et~al.(2025{\natexlab{a}})Zhang, Li, Qi, Li, Liu, Wang, Liu, Zhou, Wu, Li, et~al.]{zhang2025embodied}
Jiazhao Zhang, Anqi Li, Yunpeng Qi, Minghan Li, Jiahang Liu, Shaoan Wang, Haoran Liu, Gengze Zhou, Yuze Wu, Xingxing Li, et~al.
\newblock Embodied navigation foundation model.
\newblock \emph{arXiv preprint arXiv:2509.12129}, 2025{\natexlab{a}}.

\bibitem[Zhang et~al.(2025{\natexlab{b}})Zhang, Wang, Wang, Li, Liu, Wei, Wang, Zhang, and Wang]{zhang2024uni}
Jiazhao Zhang, Kunyu Wang, Shaoan Wang, Minghan Li, Haoran Liu, Songlin Wei, Zhongyuan Wang, Zhizheng Zhang, and He~Wang.
\newblock Uni-navid: A video-based vision-language-action model for unifying embodied navigation tasks.
\newblock \emph{Robotics: Science and Systems}, 2025{\natexlab{b}}.

\bibitem[Zhang et~al.(2025{\natexlab{c}})Zhang, Hao, Tang, Fu, Zheng, Wang, Wang, Ding, and Zhang]{zhang2025nava}
Lingfeng Zhang, Xiaoshuai Hao, Yingbo Tang, Haoxiang Fu, Xinyu Zheng, Pengwei Wang, Zhongyuan Wang, Wenbo Ding, and Shanghang Zhang.
\newblock Nava$^3$: Understanding any instruction, navigating anywhere, finding anything.
\newblock \emph{arXiv preprint arXiv:2508.04598}, 2025{\natexlab{c}}.

\bibitem[Zhang et~al.(2025{\natexlab{d}})Zhang, Hao, Xu, Zhang, Zhang, Wang, Zhang, Wang, Zhang, and Xu]{zhang-etal-2025-mapnav}
Lingfeng Zhang, Xiaoshuai Hao, Qinwen Xu, Qiang Zhang, Xinyao Zhang, Pengwei Wang, Jing Zhang, Zhongyuan Wang, Shanghang Zhang, and Renjing Xu.
\newblock {M}ap{N}av: A novel memory representation via annotated semantic maps for {VLM}-based vision-and-language navigation.
\newblock In \emph{Proceedings of the 63rd Annual Meeting of the Association for Computational Linguistics}, pages 13032--13056, Vienna, Austria, July 2025{\natexlab{d}}.

\bibitem[Zhang et~al.(2025{\natexlab{e}})Zhang, Liu, Zhang, Aghaei, Hu, Gu, Alomrani, Bravo, Karimi, Hamidizadeh, et~al.]{zhang2025mem2ego}
Lingfeng Zhang, Yuecheng Liu, Zhanguang Zhang, Matin Aghaei, Yaochen Hu, Hongjian Gu, Mohammad~Ali Alomrani, David Gamaliel~Arcos Bravo, Raika Karimi, Atia Hamidizadeh, et~al.
\newblock Mem2ego: Empowering vision-language models with global-to-ego memory for long-horizon embodied navigation.
\newblock \emph{arXiv preprint arXiv:2502.14254}, 2025{\natexlab{e}}.

\bibitem[Zhang et~al.(2025{\natexlab{f}})Zhang, Du, Wu, Zhou, Qi, Ma, and Zhou]{zhang2025apexnav}
Mingjie Zhang, Yuheng Du, Chengkai Wu, Jinni Zhou, Zhenchao Qi, Jun Ma, and Boyu Zhou.
\newblock Apexnav: An adaptive exploration strategy for zero-shot object navigation with target-centric semantic fusion.
\newblock \emph{arXiv preprint arXiv:2504.14478}, 2025{\natexlab{f}}.

\bibitem[Zhang et~al.(2025{\natexlab{g}})Zhang, Yu, Su, and Wang]{zhang2025reinflow}
Tonghe Zhang, Chao Yu, Sichang Su, and Yu~Wang.
\newblock Reinflow: Fine-tuning flow matching policy with online reinforcement learning.
\newblock \emph{arXiv preprint arXiv:2505.22094}, 2025{\natexlab{g}}.

\bibitem[Zhang et~al.(2024{\natexlab{b}})Zhang, Wu, Li, Li, Ma, Liu, and Li]{zhang2024video}
Yuanhan Zhang, Jinming Wu, Wei Li, Bo~Li, Zejun Ma, Ziwei Liu, and Chunyuan Li.
\newblock Video instruction tuning with synthetic data.
\newblock \emph{arXiv preprint arXiv:2410.02713}, 2024{\natexlab{b}}.

\bibitem[Zhang et~al.(2025{\natexlab{h}})Zhang, Zhu, Pan, Wang, Xu, Sun, and Zheng]{zhang2025activevln}
Zekai Zhang, Weiye Zhu, Hewei Pan, Xiangchen Wang, Rongtao Xu, Xing Sun, and Feng Zheng.
\newblock Activevln: Towards active exploration via multi-turn rl in vision-and-language navigation.
\newblock \emph{arXiv preprint arXiv:2509.12618}, 2025{\natexlab{h}}.

\bibitem[Zhao et~al.(2025)Zhao, Lu, Kim, Fu, Zhang, Wu, Li, Ma, Han, Finn, et~al.]{zhao2025cot}
Qingqing Zhao, Yao Lu, Moo~Jin Kim, Zipeng Fu, Zhuoyang Zhang, Yecheng Wu, Zhaoshuo Li, Qianli Ma, Song Han, Chelsea Finn, et~al.
\newblock Cot-vla: Visual chain-of-thought reasoning for vision-language-action models.
\newblock In \emph{Proceedings of the Computer Vision and Pattern Recognition Conference}, pages 1702--1713, 2025.

\bibitem[Zhong et~al.(2024)Zhong, Wu, Ci, Wang, and Chen]{zhong2024empowering}
Fangwei Zhong, Kui Wu, Hai Ci, Churan Wang, and Hao Chen.
\newblock Empowering embodied visual tracking with visual foundation models and offline rl.
\newblock In \emph{European Conference on Computer Vision}, pages 139--155. Springer, 2024.

\bibitem[Zhou et~al.(2024)Zhou, Hong, and Wu]{zhou2023navgpt}
Gengze Zhou, Yicong Hong, and Qi~Wu.
\newblock Navgpt: Explicit reasoning in vision-and-language navigation with large language models.
\newblock In \emph{Proceedings of the AAAI Conference on Artificial Intelligence}, volume~38, pages 7641--7649, 2024.

\bibitem[Zhou et~al.(2025)Zhou, Hong, Wang, Wang, and Wu]{zhou2025navgpt}
Gengze Zhou, Yicong Hong, Zun Wang, Xin~Eric Wang, and Qi~Wu.
\newblock Navgpt-2: Unleashing navigational reasoning capability for large vision-language models.
\newblock In \emph{European Conference on Computer Vision}, pages 260--278, 2025.

\bibitem[Zhou et~al.()Zhou, Zhu, Liu, Tang, Wen, Peng, Shen, and Xu]{zhouchatvla}
Zhongyi Zhou, Yichen Zhu, Xiaoyu Liu, Zhibin Tang, Junjie Wen, Yaxin Peng, Chaomin Shen, and Yi~Xu.
\newblock Chatvla-2: Vision-language-action model with open-world reasoning.
\newblock In \emph{The Thirty-ninth Annual Conference on Neural Information Processing Systems}.

\bibitem[Zhu et~al.(2021)Zhu, Liang, Zhu, Yu, Chang, and Liang]{zhu2021soon}
Fengda Zhu, Xiwen Liang, Yi~Zhu, Qizhi Yu, Xiaojun Chang, and Xiaodan Liang.
\newblock Soon: Scenario oriented object navigation with graph-based exploration.
\newblock In \emph{Proceedings of the IEEE/CVF Conference on Computer Vision and Pattern Recognition}, pages 12689--12699, 2021.

\bibitem[Zhu et~al.(2025)Zhu, Wang, Li, Zhang, Ma, Chen, Jia, Liang, Yu, Deng, et~al.]{zhu2025mtu}
Ziyu Zhu, Xilin Wang, Yixuan Li, Zhuofan Zhang, Xiaojian Ma, Yixin Chen, Baoxiong Jia, Wei Liang, Qian Yu, Zhidong Deng, et~al.
\newblock Move to understand a 3d scene: Bridging visual grounding and exploration for efficient and versatile embodied navigation.
\newblock In \emph{Proceedings of the IEEE/CVF International Conference on Computer Vision}, pages 8120--8132, 2025.

\bibitem[Ziliotto et~al.(2025)Ziliotto, Campari, Serafini, and Ballan]{ziliotto2025tango}
Filippo Ziliotto, Tommaso Campari, Luciano Serafini, and Lamberto Ballan.
\newblock Tango: training-free embodied ai agents for open-world tasks.
\newblock In \emph{Proceedings of the Computer Vision and Pattern Recognition Conference}, pages 24603--24613, 2025.

\end{thebibliography}

\let\cleardoublepage\clearpage
\pagestyle{empty}


\end{document}